\documentclass{article}

\usepackage{microtype}
\usepackage{graphicx}
\usepackage{booktabs} % for professional tables

\usepackage{hyperref}

\usepackage[accepted]{icml2022}

\usepackage{amsmath}
\usepackage{amssymb}
\usepackage{mathtools}
\usepackage{amsthm}
\usepackage{amsfonts,bm}
\usepackage{algorithm}
\usepackage{algorithmic}
\usepackage{subcaption}
\usepackage{multirow}

\usepackage[capitalize,noabbrev]{cleveref}

\theoremstyle{plain}
\newtheorem{theorem}{Theorem}[section]
\newtheorem{proposition}[theorem]{Proposition}

\theoremstyle{definition}
\newtheorem{definition}[theorem]{Definition}

\theoremstyle{remark}

\newenvironment{myproof}{{\noindent\it Proof.}\noindent}{\hfill $\square$\par}

\usepackage[textsize=tiny]{todonotes}

\icmltitlerunning{Structural Entropy Guided Graph Hierarchical Pooling}

\begin{document}

\twocolumn[
\icmltitle{Structural Entropy Guided Graph Hierarchical Pooling}



\icmlsetsymbol{equal}{*}

\begin{icmlauthorlist}
\icmlauthor{Junran Wu}{nlsde,equal}
\icmlauthor{Xueyuan Chen}{nlsde,equal}
\icmlauthor{Ke Xu}{nlsde}
\icmlauthor{Shangzhe Li}{nlsde,math}
\end{icmlauthorlist}

\icmlaffiliation{nlsde}{State Key Lab of Software Development Environment, Beihang University, Beijing, 100191, China}
\icmlaffiliation{math}{School of Mathematical Science, Beihang University, Beijing, 100191, China}

\icmlcorrespondingauthor{Shangzhe Li}{shangzheli@buaa.edu.cn}

\icmlkeywords{Structural Entropy, Hierarchical Pooling, Graph Classification, Node Classification}

\vskip 0.3in
]



\printAffiliationsAndNotice{\icmlEqualContribution} % otherwise use the standard text.

\begin{abstract}
Following the success of convolution on non-Euclidean space, the corresponding pooling approaches have also been validated on various tasks regarding graphs. However, because of the fixed compression quota and stepwise pooling design, these hierarchical pooling methods still suffer from local structure damage and suboptimal problem.
In this work, inspired by structural entropy, we propose a hierarchical pooling approach, SEP, to tackle the two issues. Specifically, without assigning the layer-specific compression quota, a global optimization algorithm is designed to generate the cluster assignment matrices for pooling at once.
Then, we present an illustration of the local structure damage from previous methods in the reconstruction of ring and grid synthetic graphs.
In addition to SEP, we further design two classification models, SEP-G and SEP-N for graph classification and node classification, respectively. The results show that SEP outperforms state-of-the-art graph pooling methods on graph classification benchmarks and obtains superior performance on node classifications.
\end{abstract}

\section{Introduction}
Chasing the great success of deep learning in natural language processing and images, plenty of research efforts have been devoted to the adoption of neural networks in tasks without data on the Euclidean domain, i.e., in graphs \cite{kipf2017semi,velivckovic2018graph}. 
Thus, recent years, graph neural networks (GNNs) become ubiquitous within deep learning for graphs, and have obtained great accomplishments in various domains, such as node classification \cite{kipf2017semi}, link prediction \cite{zhang2018link} and graph classification \cite{xu2019powerful}. In these works, a key direction is the convolutional mechanism of GNNs \cite{wu2019simplifying,zhu2020simple,wu2022simple}, which aims to learn the structural information in graphs. Meanwhile, another research direction in GNNs is the pooling mechanism, which follows the customs in CNN models that compress a set of nodes into a compact representation \cite{lee2019self,bianchi2020spectral}.

\begin{figure}[!t]
\vskip 0.2in
\begin{center}
\centerline{\includegraphics[width=\columnwidth]{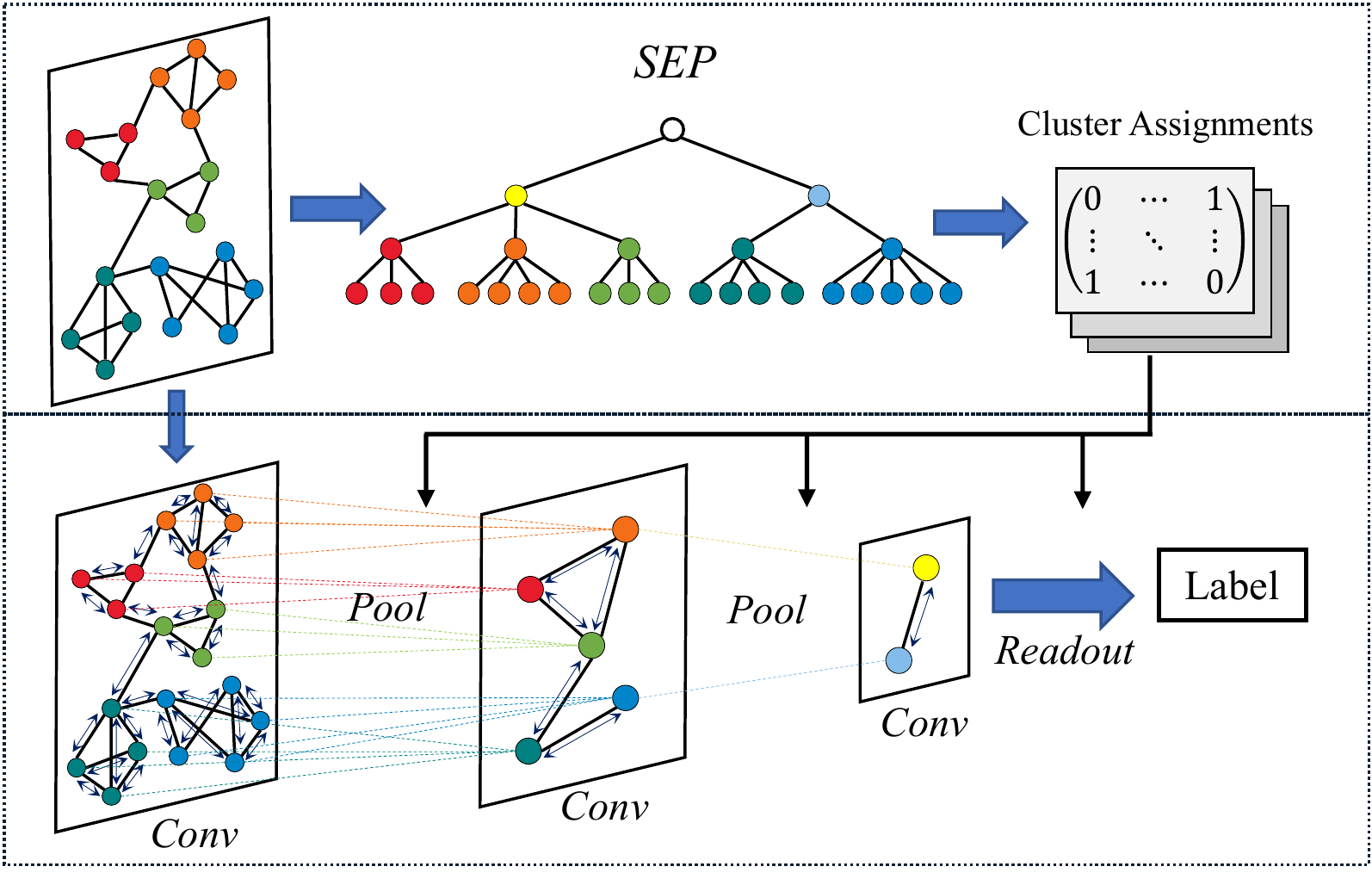}}
\caption{\textbf{Architecture of proposed SEP operator combined with graph neural network.} The graph neural network consists of message passing layers and hierarchical pooling layers. A separate algorithm is proposed for cluster assignments generation.} 
\label{fig:sep-framework}
\end{center}
\vskip -0.2in
\end{figure}

Besides the simplest pooling methods, that sum or average all nodes, various well-designed pooling approaches have been proposed to aggregate the node information in a hierarchical manner.
However, despite the effectiveness of these methods on various tasks, there are still many issues that hinder the development of GNNs.
First, pooling methods based on node drop, like TopKPool \cite{gao2019graph}, SAGPool \cite{lee2019self} and ASAP \cite{ranjan2020asap}, unnecessarily cut nodes based on designed ranking strategy in each pooling layer, resulting in information loss \cite{baek2021accurate}. Although other methods based on node clustering avoid this issue \cite{ying2018hierarchical,bianchi2020spectral}, the damage on the graph local structure still can not be prevented due to the artificially specified node compression quota \cite{gao2019graph,ranjan2020asap}, which also exists in the node drop methods \footnote{The clustering-based pooling methods require the fixed number of clusters and the node drop methods require the fixed node compression ratio.}.
Furthermore, the cluster assignments produced by these works only rely on the topology of graph in the current layer without any consideration of the relationships among pooling layers. This would probably lead to suboptimal results in their tasks.
Thus, a pooling operation that is globally optimized and has natural node partition is preferred.

In this paper, we present a hierarchical pooling method, termed \textit{SEP}, to address the above two issues that hinder the development of GNNs (Figure~\ref{fig:sep-framework}). Specifically, inspired by structural entropy \cite{li2016structural}, a metric designed to assess the graph structural information, the essential structure of a graph can be decoded by this metric as a measure of the complexity of its hierarchical structure. 
In particular, the cluster assignments for hierarchical pooling can be directly obtained from the proposed algorithm for structural entropy minimization. 
Note that, the proposed algorithm is globally optimized and free from learning, which means the cluster assignments will be generated together to avoid the suboptimal problem. Moreover, the algorithm does not rely on a fixed layer-specific compression quota but the number of compression layers, which would help retain the local structure of graphs.

Before the validation of SEP on classification benchmarks, we first present an illustration of the damage caused by previous hierarchical pooling methods on local structure of graphs. 
With seven benchmarks from bioinformatics and social networks, we then experimentally validate the effectiveness of SEP on tasks regarding graph classification, and conclude that SEP surpasses the state-of-the-art (SOTA) hierarchical pooling methods on most benchmarks especially on the social network datasets.
Besides the superior performance on global attribute discerning, we further evaluate SEP on node classification tasks to better validate its capability of information retaining within the process of pooling. The results show that SEP outperforms the model with the same architecture (i.e., g-U-Nets) and most baselines.
To sum up, our contributions are listed as follows:

\begin{itemize}
    \item We uncover two crucial issues in previous hierarchical pooling works that hinder the development of GNNs, including the local structure damage and suboptimal problem because of the fixed compression quota and stepwise pooling design.

    \item Through the introduction of the structural information theory, we present a novel hierarchical pooling approach, termed SEP, to address the unveiled issues.

    \item We extensively validate SEP on graph reconstruction, graph classification, and node classification tasks, in which outperformances are observed in comparison the SOTA hierarchical pooling methods.
\end{itemize}

\section{Related Work}
\paragraph{Hierarchical pooling.} To pursue better generalization and performance, pooling operations have been adopted to amplify the receptive fields and reduce the input sizes.
Several designs are proposed from the angle of selecting the most important $k$ nodes from the original graph  to organize a new one, such as TopKPool \cite{gao2019graph}, SAGPool \cite{lee2019self} and ASAP \cite{ranjan2020asap}. Though efficient, this node-drop design would result in information loss and isolated subgraphs, which will deteriorate the performance of GNNs.
Thus, another design based on node clustering emerges and avoids this issue, including DiffPool \cite{ying2018hierarchical} and MinCutPool \cite{bianchi2020spectral}, in which the nodes of original graph are merged into a bunch of clusters. Although preventing information loss, the damage on graph local structures still exists because of the fixed node compression quota.

\paragraph{Structural entropy.} Information entropy stems from the demand for information measurement in communication systems \cite{shannon1948mathematical}. 
Correspondingly, to measure the information in graphs, structural entropy was proposed and used to evaluate the complexity of the hierarchical structure of a graph \cite{li2016structural}, which is also a natural node clustering method for graphs. Furthermore, two- and three-dimensional structural entropy, which measure the complexity of two- and three-level hierarchical structures, respectively, have been applied in medicine \cite{li2016three}, bioinformatics \cite{li2018decoding}, and the security of networks \cite{li2016resistance}. 
In the light of this global measurement of graph information, structural entropy can be used to decode the essential structure of graphs, which further sparks us the yielding of SEP to address the two issues that impede the development of GNNs.

\section{Proposed Method}
In this section, under the guidance of structural entropy, we present our key idea and an algorithm for the cluster assignments construction. Then, we design a GNN model, which has several convolutional layers and pooling layers, to learn global representations for graph classification. Furthermore, we develop another model, which is made up of additional convolutional layers and unpooling layers, to obtain local representations for node classification. Before elaborating on them, we first show some notations.

\subsection{Preliminaries}
\label{sec:preliminaries}
A graph $G$ can be represented as a multi-tuple $(\mathcal{V}, \mathcal{E}, \mathbf{X})$, where $|\mathcal{V}|=n$ is the node set, $|\mathcal{E}| = m$ is the edge set, and $\mathbf{X}\in \mathbb{R}^{n\times d}$ is the feature matrix for $n$ nodes with $d$-dimensional feature vector. The topology structure of graph $G$ can be found in its adjacency matrix $\mathbf{A}\in \mathbb{R}^{n\times n}$.

\paragraph{Graph neural networks.} In this work, we select Graph Convolutional Network (i.e., GCN \cite{kipf2017semi}) as the convolutional layers of our models. GCN and its variants have achieved excellent performance in different kinds of tasks regarding graphs. 
There is no doubt that our proposed pooling operator can also work with other GNNs like GAT \cite{velivckovic2018graph} and GIN \cite{xu2019powerful}. This will be discussed in the experimental section.
For a stacked graph neural networks, the $i$-th convolutional layer in the form of GCN can be formally written as:
\begin{equation}
\label{eq:gcn}
H_{i+1} = ReLU(\tilde{\mathbf{D}}^{-\frac{1}{2}}\tilde{\mathbf{A}}\tilde{\mathbf{D}}^{-\frac{1}{2}}H_iW_i),
\end{equation}
where $ReLU$ is a non-linear activation function, $W_i\in\mathbb{R}^{h\times h}$ is the trainable matrix for this layer, $H_{i+1}\in\mathbb{R}^{n\times h}$ is the output of this layer and $H_0 = \mathbf{X}$. In particular, $\tilde{\mathbf{A}} = \mathbf{A}+\mathbf{I}$ denotes the adjacency matrix with self-loops, and $\tilde{\mathbf{D}}_{ii}=\sum_j\tilde{\mathbf{A}}_{ij}$.
In the process of hyper-parameter tuning, we fix the same hidden dimension for all layers.

\paragraph{Hierarchical pooling.} In general, hierarchical pooling is a graph coarsening process to dig out a subset of representative nodes and form a new graph.
Let $\mathbf{S}_{i}\in\mathbb{R}^{n_{i+1}\times n_i}$ denote the cluster assignment matrix at $i$-th pooling layer, where $n_{i}$ and $n_{i+i}$ are the number of nodes before and after graph coarsening.
The new adjacency matrix and node feature matrix after pooling are calculated by the next equations:
\begin{equation}
\label{eq:hierarchical_pooling}
\mathbf{A}_{i+1} = \mathbf{S}_i\mathbf{A}_i\mathbf{S}_i^\top;\quad \mathbf{P}_{i+1} = \mathbf{S}_iH_i,
\end{equation}
where $\mathbf{A}_{i}\in\mathbb{R}^{n_i\times n_i}$ is the adjacency matrix at the $i$-th layer and $H_i\in\mathbb{R}^{n_i\times h}$ refers to the node feature matrix produced by the $i$-th graph convolutional layer.
Specifically, according to the cluster assignments $\mathbf{S}_i$, $\mathbf{P}_{i+1}$ receives the node hidden features $H_i$ and merges these features to refine the initial representations for the $n_{i+1}$ clusters in the novel graph. Correspondingly, $\mathbf{A}_{i+1}$ employs the node connectivities $\mathbf{A}_i$ to weave a more delicate origination among $n_{i+1}$ clusters.

\subsection{Cluster Assignments via Structural Entropy Minimization}
Here, we present our methodology for hierarchical pooling. As claimed above, although plenty of works have been presented for graph coarsening based on heuristics or theories, little attention has been paid to the relationships among pooling layers. These studies only produce the cluster assignments based on the graph in the current layer.
Meanwhile, in various tasks regarding graphs, there are graphs constructed from specific domain (e.g., social networks) or weaved by human (e.g., synthetic graphs), which are usually not optimal for GNNs because of noisy information.
It is thus vital to eliminate such noisy structure from graphs in the process of cluster assignments generation.
In this paper, inspired by structural entropy \cite{li2016structural}, we propose a novel hierarchical pooling approach, denoted as SEP, to address the aforementioned issues in previous works. 
Besides the measurement of graph information, structural entropy can also be used to decode the hierarchical structure of a given graph as a metric of the complexity of its underlying essential structure. 
Thus, through structural entropy minimization, the hierarchical structure of a graph can be decoded into a corresponding coding tree, in which disturbance derived from noise or stochastic variation can be minimized \cite{li2018decoding}.
We believe an effective structural entropy minimization algorithm could uncover the connections among hierarchical pooling layers and eliminate the noisy structure in graphs. 

Based on the definition in \cite{li2016structural}, let a two-tuple $(\mathcal{V}, \mathcal{E})$ be a graph $G$, the formal equation of the structural entropy for $G$ on coding tree $T$ can be written as:
\begin{equation}
\label{eq:structural_entropy}
\mathcal{H}^T (G)=-\sum_{v_t\in T} \frac{g_{v_t}}{vol(\mathcal{V})} \log \frac{vol(v_t)}{vol(v_t^+)},
\end{equation}
where $v_t$ is a nonroot node in $T$ that can also be viewed as a node subset $\subset \mathcal{V}$ according to its leaf node partition in $T$, $v_t^+$ is the parent of $v_t$, $g_{v_t}$ refers to the number of edges with an endpoint in the leaf node partition of $v_t$, and $vol(\mathcal{V})$ and $vol(v_t)$ are the sums of degrees of leaf nodes in $\mathcal{V}$ and $v_t$, respectively. 
The minimum entropy realized by the optimal coding tree $T$ is the structural entropy of $G$, which follows this target equation: $\mathcal{H}(G)=\min_{\forall T}\{\mathcal{H}^T (G)\}.$ 
According to the definition of structural entropy, we know that the coding tree is a natural hierarchical division for graphs, and the connections among different layers are established for the purpose of structural entropy minimization. Furthermore, the local structure in graphs will be retained because we do not need to allocate layer-specific node compression quotas.

Besides the optimal coding tree that realizes the minimized structural entropy, in most cases, a natural coding tree with a certain height is preferred, because we only need a few fixed times of graph coarsening for most tasks \cite{gao2019graph,baek2021accurate}.
In this context, the $k$-dimensional structural entropy of $G$ is proposed to decode the optimal coding tree with a fixed height $k$:
\begin{equation}
\mathcal{H}^{(k)}(G)=\min_{\forall T:\text{Height}(T)=k}\{\mathcal{H}^T (G)\}.
\end{equation}
In this paper, under the guidance of $k$-dimensional structural entropy, we aim to investigate the solution for decrypting the coding tree with a certain height $k$. Firstly, we define three functions.

\begin{definition}
\label{def:merge}
Let $T$ be any coding tree for graph $G=(\mathcal{V}, \mathcal{E})$, $v_r$ is the root node of $T$ and $\mathcal{V}$ are the leaf nodes of $T$. 
Given any two nodes $(v_i, v_j)$ in $T$, in which $v_i\in v_r.children$ and $v_j\in v_r.children$. 

Define a function $\text{MERGE}_T(v_i, v_j)$ for $T$ to insert a new node $v_\varepsilon$ between $v_r$ and $(v_i, v_j)$:
\begin{align}
  v_\varepsilon.children \leftarrow v_i; \\
  v_\varepsilon.children \leftarrow v_j; \\
  v_r.children \leftarrow v_\varepsilon;
\end{align}
\end{definition}

\begin{definition}
Following the setting in Definition \ref{def:merge}, given any two nodes $(v_i, v_j)$, in which $v_i\in v_j.children$. 

Define a function $\text{REMOVE}_T(v_i)$ for $T$ to remove $v_i$ from $T$ and merge $v_i.children$ to $v_j.chileren$:
\begin{equation}
  v_j.children \leftarrow v_i.children; \\
\end{equation}
\end{definition}

\begin{definition}
Following the setting in Definition \ref{def:merge}, given any two nodes $(v_i, v_j)$, in which $v_i\in v_j.children$ and $|Heigth(v_j)-Height(v_i)|>1$. 

Define a function $\text{FILL}(v_i, v_j)$ for $T$ to insert a new node $v_\varepsilon$ between $v_i$ and $v_j$:
\begin{align}
  v_\varepsilon.children \leftarrow v_i; \\
  v_j.children \leftarrow v_\varepsilon; 
\end{align}
\end{definition}

\begin{algorithm}[!t]
\caption{Coding tree with height $k$ via structural entropy minimization}
\label{code:coding_tree} 
\textbf{Input:} a graph $G=(\mathcal{V}, \mathcal{E})$, a positive integer $k>1$\\
\textbf{Output:} a coding tree $T$ with height $k$

\begin{algorithmic}[1]
\STATE Generate a coding tree $T$ with a root node $v_r$ and all nodes in $\mathcal{V}$ as leaf nodes;
\STATE // Stage 1: Bottom to top construction;
\WHILE{$|v_r.children|>2$} {
  \STATE Select $v_i$ and $v_j$ from $v_r.children$, conditioned on \\
  $argmax_{(v_i, v_j)}\{\mathcal{H}^T(G) - \mathcal{H}^{T_{\text{MERGE}(v_i, v_j)}}(G)\}$;
  \STATE $\text{MERGE}(v_i, v_j)$;
}
\ENDWHILE
\STATE // Stage 2: Compress $T$ to the certain height $k$;
\WHILE{$\text{Height}(T)>k$} {
  \STATE Select $v_i$ from $T$, conditioned on \\
  $argmin_{v_i}\{\mathcal{H}^{T_{\text{REMOVE}(v_i)}}(G) - \mathcal{H}^T(G)|$\\
  \qquad \qquad \,\, $v_i\neq v_r \,\&\, v_i\notin \mathcal{V}\}$;
  \STATE $\text{REMOVE}(v_i)$;
}
\ENDWHILE
\STATE // Stage 3: Fill $T$ to avoid cross-layer links;
\FOR{$v_i\in T$} {
  \IF{$|\text{Height}(v_i.parent)-\text{Height}(v_i)|>1$} 
    \STATE $\text{FILL}(v_i, v_i.parent)$;
  \ENDIF
}
\ENDFOR
\STATE return $T$;
\end{algorithmic} 
\end{algorithm}

Based on the three defined functions, a greedy algorithm to compute the coding tree with a certain height $k$ via structural entropy minimization can be found in Algorithm \ref{code:coding_tree}.
Specifically, a full-height binary coding tree is first generated from bottom to top. In this stage, two child nodes of root are merged to form a new division in each iteration, which aims to maximize the structural entropy reduction.
In the second stage, in order to satisfy a fixed number of graph coarsening, we need squeeze the previous full-height binary coding tree by dropping nodes. Each time, we select an inner-node from $T$, which makes $T$ have the minimized structural entropy after removing this node. 
At the end of the second stage, we have already obtained a coding tree with a certain height $k$ under the guidance of structural entropy. However, there may be nodes that do not have immediate successor in the next layer because of cross-layer links, which will cause node missing when realizing hierarchical pooling based on such a coding tree. Therefore, we need perform the third stage to ensure the integrity of information transmission between layers, and also need not interfere with the structural entropy of $G$ on coding tree $T$ (see Proposition \ref{prop:ct_fill}).
Finally, a coding tree $T$ for the given graph $G$ can be obtained, where $T=(\mathcal{V}^T, \mathcal{E}^T)$, $\mathcal{V}^T=(\mathcal{V}^T_0, \dots, \mathcal{V}^T_k)$ and $\mathcal{V}^T_0=\mathcal{V}$. In addition, the cluster assignment matrices can also be obtained from $\mathcal{E}^T$, that is, $\mathbb{S} = (\mathbf{S}_1, \dots, \mathbf{S}_k)$.

\begin{proposition}
\label{prop:ct_fill}
Let $T$ be a coding tree after the second stage of Algorithm \ref{code:coding_tree}, and given two adjacent nodes $(v_i, v_j)$ in $T$, in which $v_i\in v_j.children$ and $|Heigth(v_j)-Height(v_i)|>1$. Then, $\mathcal{H}^T(G) = \mathcal{H}^{T_{\text{FILL}(v_i, v_j)}}(G)$.
\end{proposition}

\begin{myproof}
Equation \ref{eq:structural_entropy} shows that the structural entropy of graph $G$ on $T$ is the summation of $\mathcal{H}^T_v=-\frac{g_v}{vol(\mathcal{V})} \log \frac{vol(v)}{vol(v^+)}$ for all nonroot nodes in $T$. That is, $\mathcal{H}^T(G)=\mathcal{H}^T_{v_i}+\mathcal{H}^T_{v_j}+\dots$ and $\mathcal{H}^{T_F}(G)=\mathcal{H}^{T_F}_{v'_i}+\mathcal{H}^{T_F}_{v_\varepsilon}+\mathcal{H}^{T_F}_{v'_j}+\dots$, denote $T_F=T_{\text{FILL}(v_i, v_j)}$ for simplicity and $(v'_i, v'_j)$ corresponds to $(v_i, v_j)$ after FILL. According to Equation \ref{eq:structural_entropy}, we have:
\begin{align}
\mathcal{H}^{T_F}_{v'_i} & = -\frac{g_{v'_i}}{vol(\mathcal{V})} \log \frac{vol(v'_i)}{vol(v'^+_i)} \nonumber \\
 & = -\frac{g_{v'_i}}{vol(\mathcal{V})} \log \frac{vol(v'_i)}{vol(v_\varepsilon)} \nonumber \\
 & =0 \quad with \quad vol(v'_i)= vol(v_\varepsilon),  \\
\mathcal{H}^{T_F}_{v_\varepsilon} & = -\frac{g_{v_\varepsilon}}{vol(\mathcal{V})} \log \frac{vol(v_\varepsilon)}{vol(v^+_\varepsilon)} \nonumber \\
& = -\frac{g_{v_i}}{vol(\mathcal{V})} \log \frac{vol(v_i)}{vol(v_j)} \nonumber \\
& = \mathcal{H}^T_{v_i}, \\
\mathcal{H}^{T_F}_{v'_j} & = -\frac{g_{v'_j}}{vol(\mathcal{V})} \log \frac{vol(v'_j)}{vol(v'^+_j)} \nonumber \\
 & = -\frac{g_{v_j}}{vol(\mathcal{V})} \log \frac{vol(v_j)}{vol(v^+_j)} \nonumber \\
 & =\mathcal{H}^T_{v_j}.
\end{align}
Thus, we have $\mathcal{H}^T(G) = \mathcal{H}^{T_{\text{FILL}(v_i, v_j)}}(G)$.
\end{myproof}

\paragraph{Complexity analysis.} The runtime complexity of Algorithm \ref{code:coding_tree} is $O(2n+h_{max}(m\log n+n))$, and $h_{max}$ is the height of coding tree $T$ after the first stage. Meanwhile, since coding tree $T$ tends to be balanced in the process of structural entropy minimization, $h_{max}$ will be around $\log n$. Furthermore, graph generally has more edges than nodes, i.e., $m\gg n$, thus the runtime of Algorithm \ref{code:coding_tree} almost scales linearly in the number of edges.

\subsection{Graph Neural Network for Graph Classification}
\begin{figure}[!h]
\vskip 0.2in
\begin{center}
\centerline{\includegraphics[width=\columnwidth]{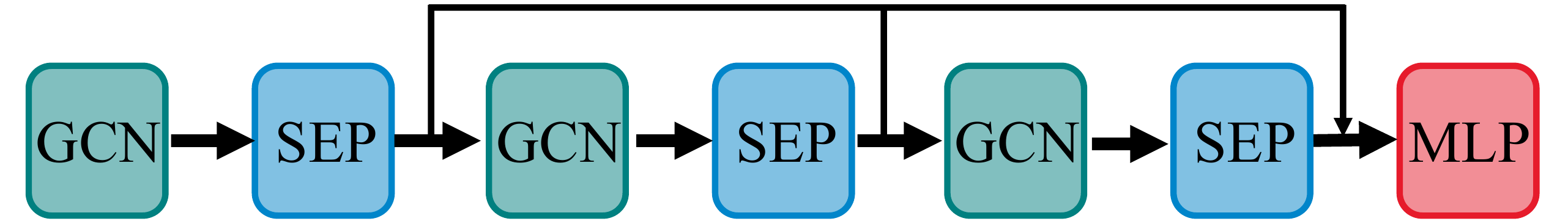}}
\caption{\textbf{The SEP-G architecture for graph classification.} Following the design of previous works in hierarchical pooling, the architecture is comprised of three GCN layers and each is followed by corresponding SEP layer.} 
\label{fig:sep-gcn}
\end{center}
\vskip -0.2in
\end{figure}

In this subsection, we present the architecture based on SEP for graph classification, and name it \textit{SEP-G}. As shown in Figure~\ref{fig:sep-gcn}, the hierarchical pooling architecture follows the setting in previous pooling studies \cite{lee2019self,bianchi2020spectral,baek2021accurate}, which consists of three blocks and each block has a GCN layer and a SEP layer. For the graph-level representation, the outputs after each block are summarized and then fed to a prediction layer for classification. Based on Equations \ref{eq:gcn} and \ref{eq:hierarchical_pooling}, the graph representation with SET-G can be formally written as:
\begin{align}
h_G = Concat(&Readout(SEP_i(GCN_i(H_i, \mathbf{A}_i), \mathbf{S}_i)) \nonumber \\
& | \forall i\in\{1, 2, 3\}).
\end{align}

\paragraph{Permutation invariance.} In graph classification, it is important to ensure the permutation invariance of designed graph neural network. In our proposed model for graph classification, there are two main components, that is, GCN layer and SEP layer. The permutation invariance of GCN layer has been confirmed by previous works \cite{ying2018hierarchical,ma2019graph}. Thus, the SEP layer should be invariant with permutations.

\begin{proposition}
Given a permutation matrix $\mathcal{P}\in\{0, 1\}^{n\times n}$, then $\text{SEP}(A, H) = \text{SEP}(\mathcal{P}A\mathcal{P}^\top, \mathcal{P}H)$ (i.e., SEP is permutation invariant).
\end{proposition}
\begin{myproof}
The cluster assignments are derived from the coding tree via Algorithm \ref{code:coding_tree}, which is a traversal algorithm that does not depend on the order of nodes. Therefore, the generated assignments $\mathbb{S}$ will not change with any permutation. In addition, we know that the permutation matrix is orthogonal, thus $\mathcal{P}\mathcal{P}^\top=I$ with Equation \ref{eq:hierarchical_pooling} finishes the proof.
\end{myproof}

\subsection{Graph Neural Network for Node Classification}
\begin{figure}[!ht]
\vskip 0.2in
\begin{center}
\centerline{\includegraphics[width=\columnwidth]{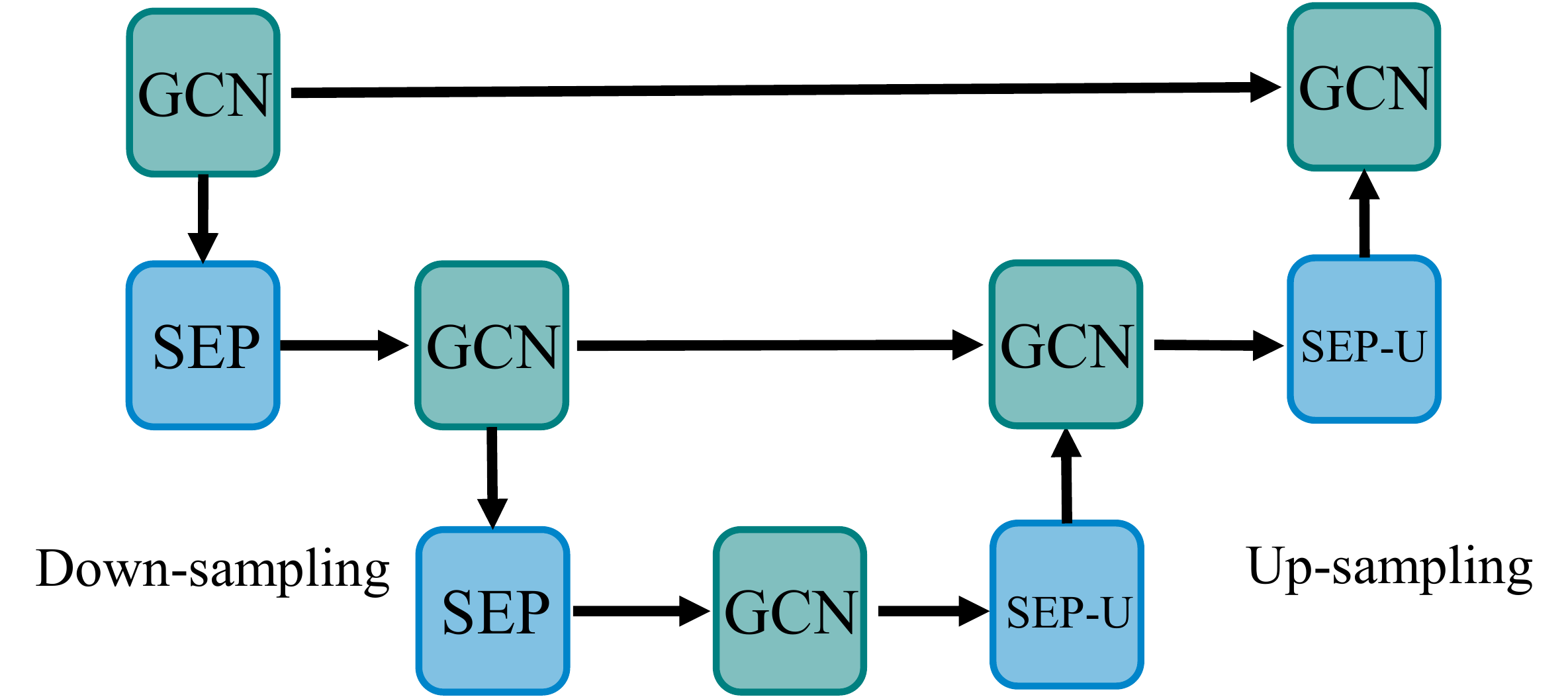}}
\caption{\textbf{The SEP-N architecture for node classification.} There are two encoder and two decoder blocks and each block is composed of a GCN layer and a pooling (unpooling) layer. Skip connection links the same-level encoder and decoder to enhance spatial feature transmission.} 
\label{fig:sep-unet}
\end{center}
\vskip -0.2in
\end{figure}

Beyond the functionality of cluster assignments to convert graphs into high-level representations, we can also adopt the same matrix $\mathbf{S}_i$ to unpool the compressed graph representation $H_i$ and structure $A_i$ to the original space by:
\begin{equation}
\mathbf{A}_{i+1} = \mathbf{S}_i^\top\mathbf{A}_i\mathbf{S}_i;\quad \mathbf{P}_{i+1} = \mathbf{S}_i^\top H_i.
\end{equation}
In this context, we present an illustration of the architecture for node classification in Figure~\ref{fig:sep-unet}, and we call it \textit{SEP-N}. SEP-N is an encoder-decoder network analogous to the design of g-U-nets \cite{gao2019graph}. In the encoder, several down-sampling blocks are applied to encode higher-order features. Each block consists of a GCN layer and a SEP layer like our SEP-G. In the decoder, we employ the consistent number of decoding components. Thus, we can see the same GCN layer but with a SEP-U layer for unpooling in the decoder block. The skip-connections linking corresponding encoders and decoders are also adopted to enable spatial information transmission for better performance.
Finally, a GCN layer is used to perform final predictions.

\section{Experiments}
\label{sec:experiments}
In this section, we describe the experiment setup for graph classification and node classification in detail, and validate the effectiveness of our proposed methods, SEP-G and SEP-N, in several corresponding benchmarks \footnote{The implementation of Algorithm \ref{code:coding_tree}, SEP-G and SEP-N can be found at \url{https://github.com/Wu-Junran/SEP}.}.

\subsection{Graph Reconstruction}
Before the presentation of experiments regarding two major classification tasks in GNNs, we first employ a graph reconstruction experiment, which quantifies the structural information retained by pooling layer, to directly reveal the damage caused by previous hierarchical pooling methods to graph's local structures.

\paragraph{Configuration.} An autoencoder is trained to reconstruct the input graph with pooling and unpooling layers. The learning objective is to minimize the mean square error (MSE) between input features $\mathbf{X}$ and output features $\mathbf{X}^r$, i.e., $min\|\mathbf{X}-\mathbf{X}^r\|^2$. For configuration, we employ the Synthetic graphs \cite{bianchi2020spectral}, including a ring and a grid that the input features are the coordinates of nodes in a 2-D Euclidean space. Detailed experiment configuration and model description can be found in Appendix A.1.

\begin{figure}[!hp]
  \vskip 0.2in
  \centering
  \begin{subfigure}{.1\textwidth}
    \centering
    \includegraphics[width=\linewidth]{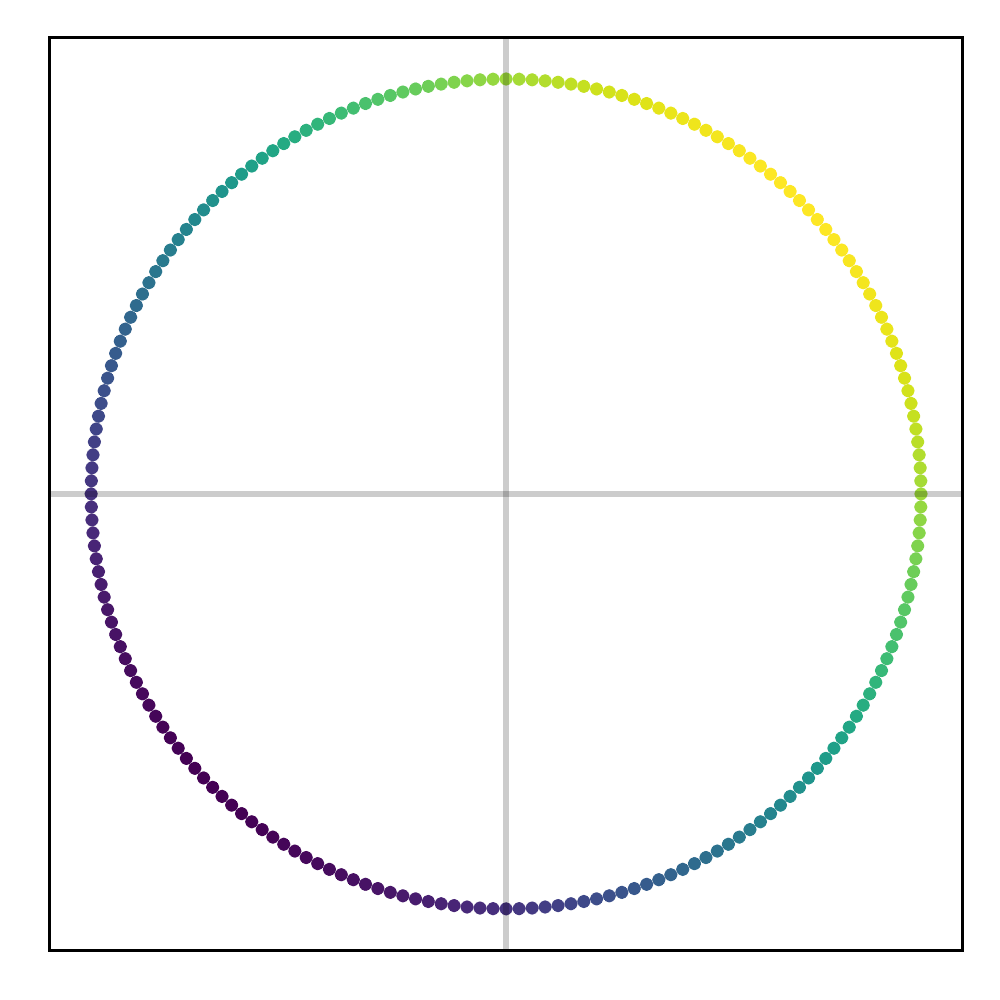}
  \end{subfigure}
  \begin{subfigure}{.1\textwidth}
    \centering
    \includegraphics[width=\linewidth]{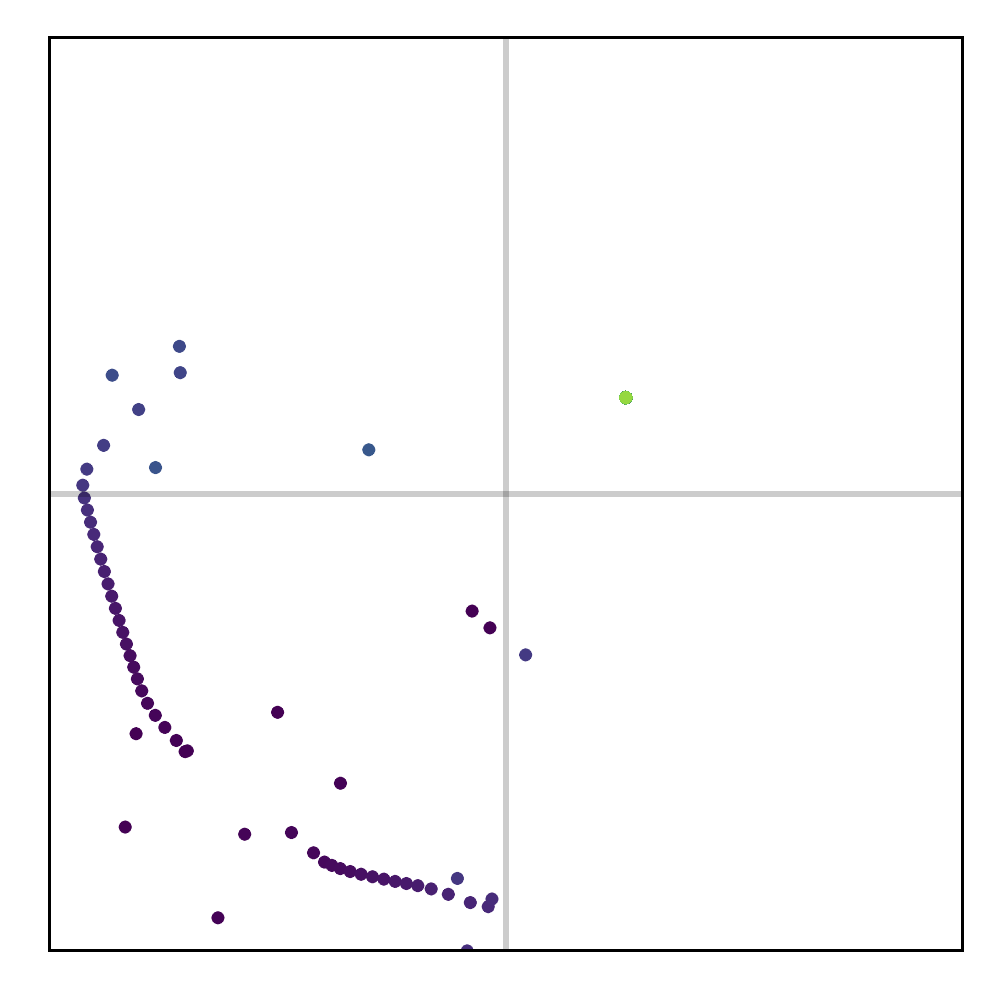}
  \end{subfigure} 
  \begin{subfigure}{.1\textwidth}
    \centering
    \includegraphics[width=\linewidth]{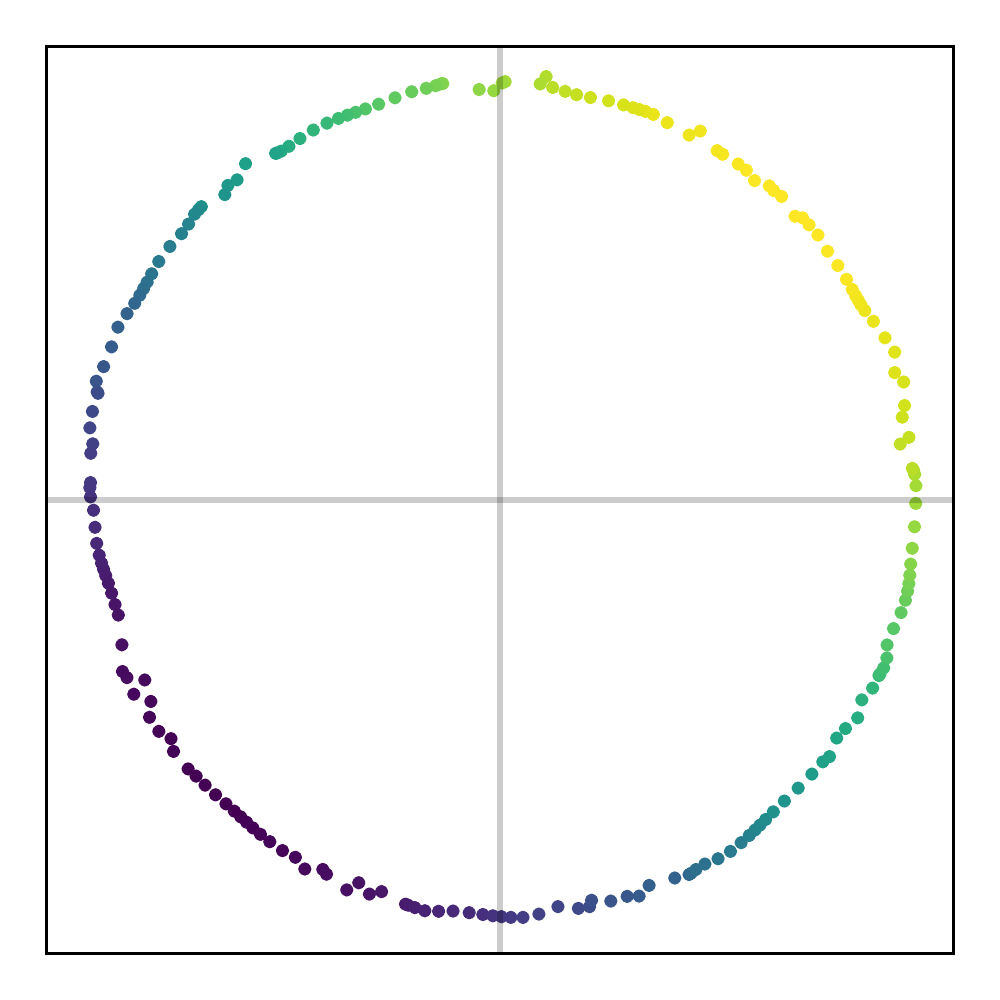}
  \end{subfigure} 
  \begin{subfigure}{.1\textwidth}
    \centering
    \includegraphics[width=\linewidth]{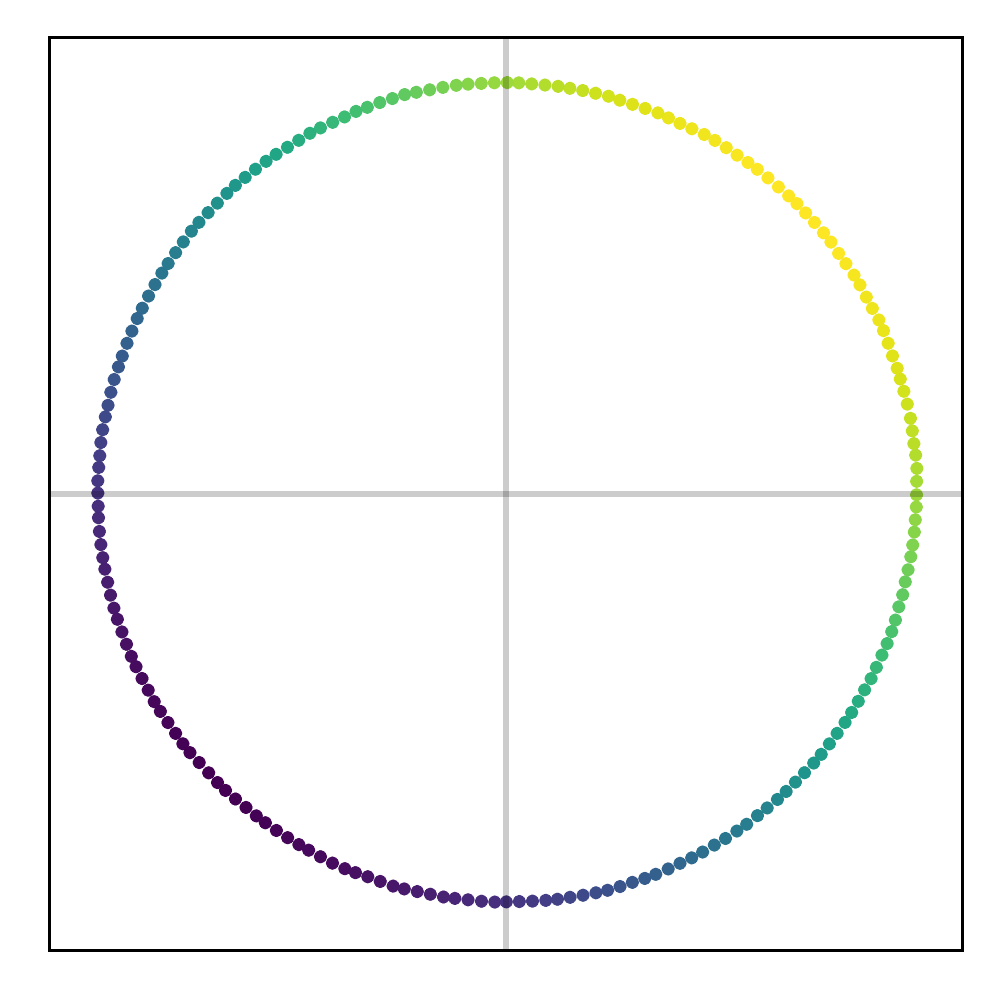}
  \end{subfigure} 
  \\
  \begin{subfigure}{.1\textwidth}
    \centering
    \includegraphics[width=\linewidth]{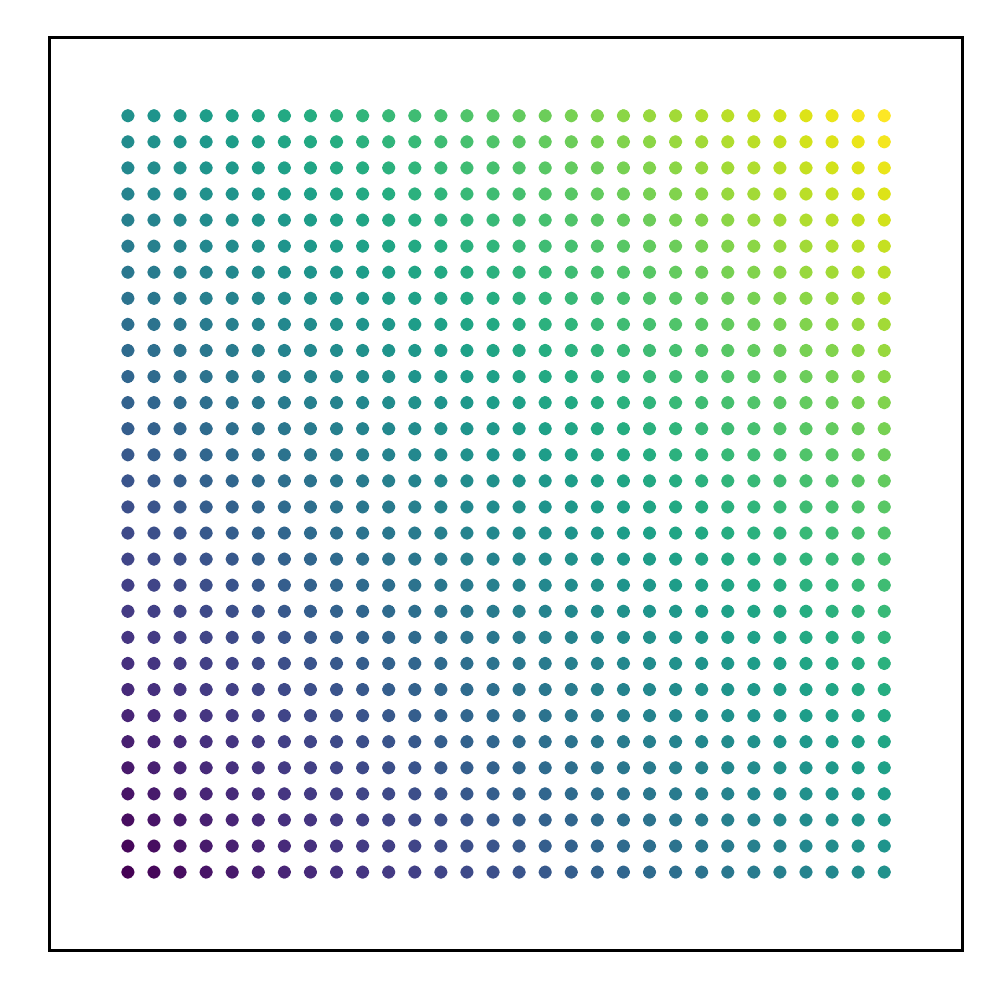}
    \caption{\tiny{Original}}
  \end{subfigure}
  \begin{subfigure}{.1\textwidth}
    \centering
    \includegraphics[width=\linewidth]{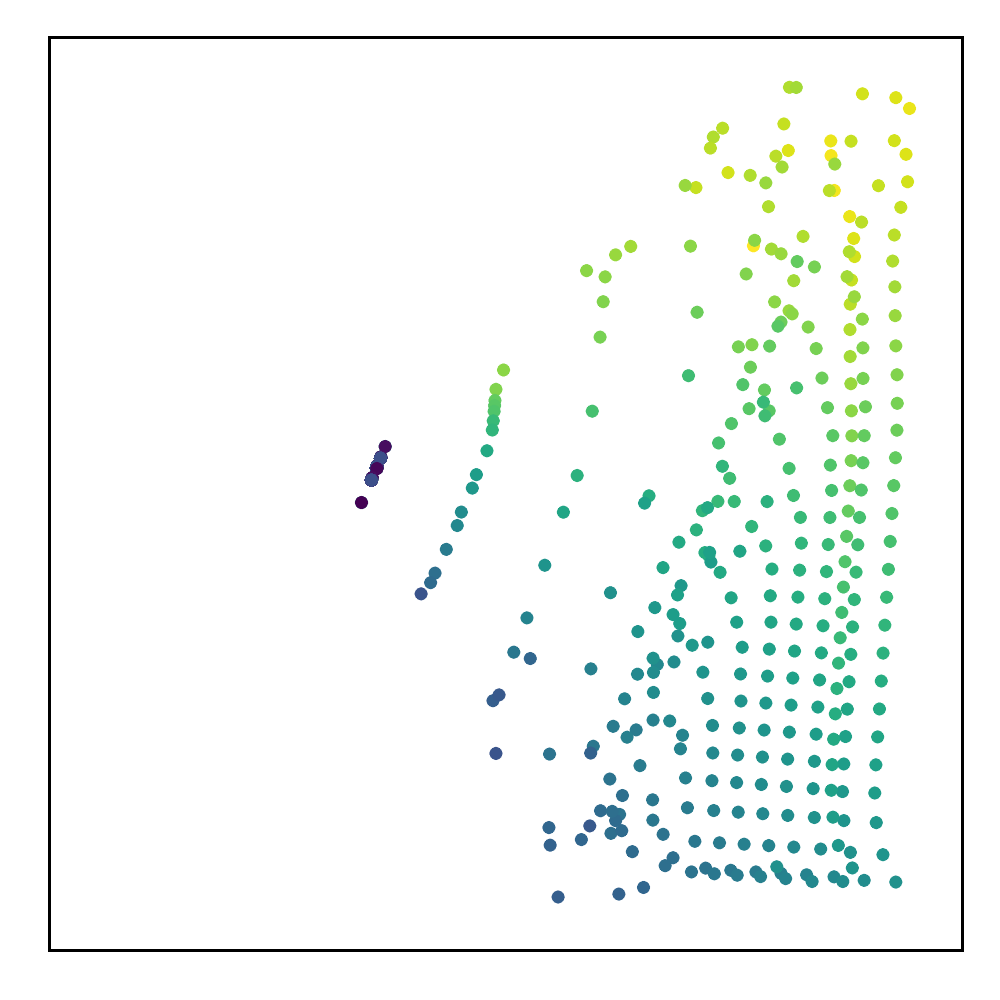}
    \caption{\tiny{TopKPool}}
  \end{subfigure} 
  \begin{subfigure}{.1\textwidth}
    \centering
    \includegraphics[width=\linewidth]{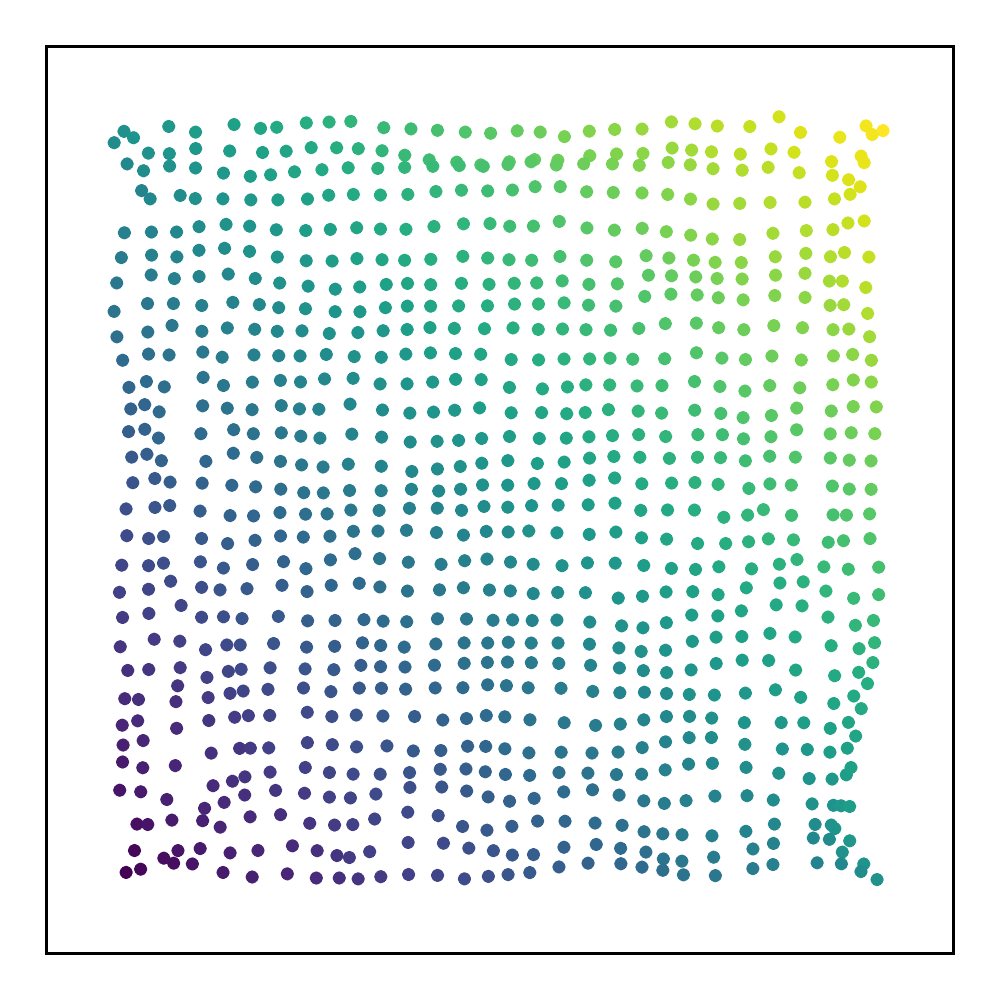}
    \caption{\tiny{MinCutPool}}
  \end{subfigure} 
  \begin{subfigure}{.1\textwidth}
    \centering
    \includegraphics[width=\linewidth]{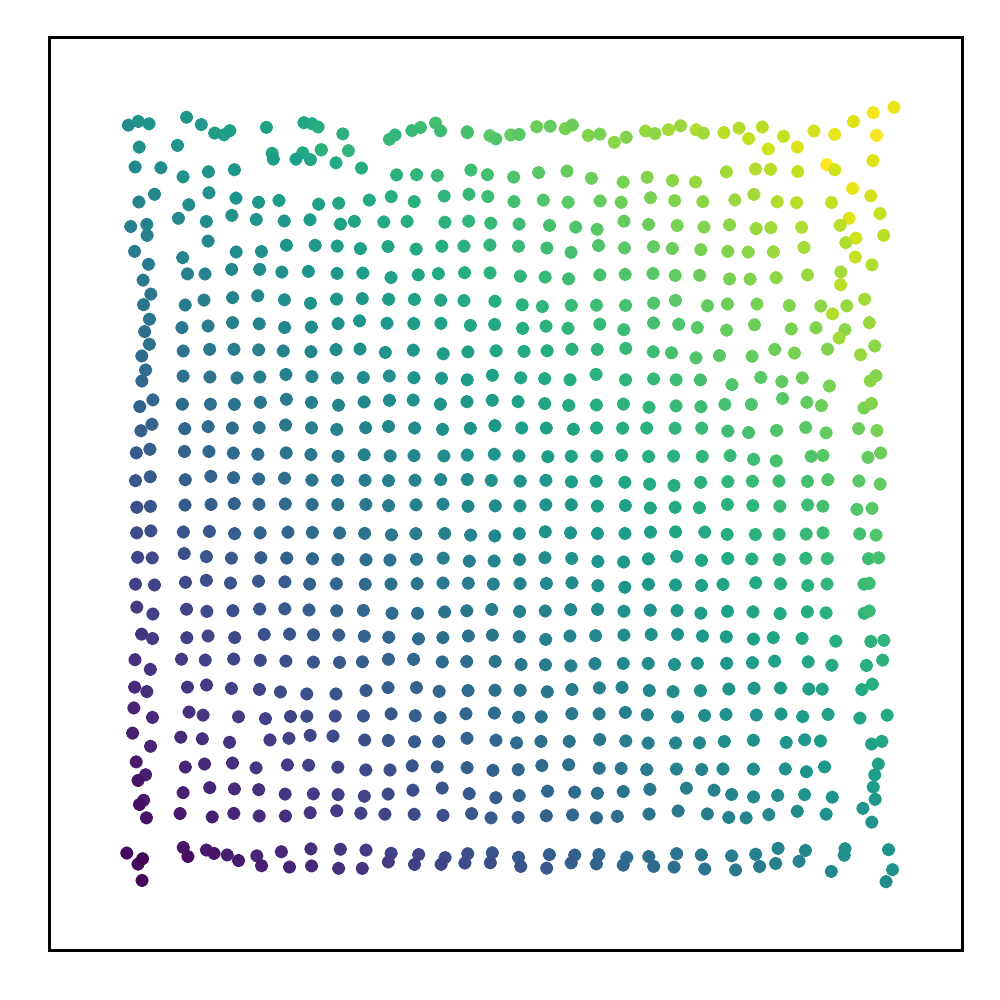}
    \caption{\tiny{SEP}}
  \end{subfigure} 
  \caption{Illustration of local structure damage from node drop and clustering methods in reconstruction of ring and grid synthetic graphs.} 
  \label{fig:resconstruction}
  \vskip 0.2in
\end{figure}

\paragraph{Reconstruction results.} Figure~\ref{fig:resconstruction} shows the original graphs and the reconstructed graphs by model with various pooling methods. We select TopKPool to represent methods with node drop design and MinCutPool to represent methods with node clustering design \footnote{Reconstruction results of all hierarchical pooling baselines can be found in Appendix A.1.}. We first notice the reconstructed results of TopKPool, where the basic shape of original graphs can not even be identified. This confirms that node drop methods will lead to severe information missing and implies the poor performance of node drop methods in graph classification. On the other hand, MinCutPool indeed retains the basic shape of original graphs. However, we can still see the significant distortion in the edge of ring and the center of grid, which represent the key structures of ring and grid, and this validates the assumption that the local structure in graphs would be ruined by the artificially specified node compression quota. 
On the contrary, SEP almost reconstructs the ring and retains the essential structure in grid center which suggests that our pooling method obtains the key structural information of original graphs.

\begin{table*}[!ht]
\caption{\textbf{Graph classification accuracies on seven benchmarks (\%).} The shown accuracies are mean and standard deviation over 10 different runs. We highlight the best results.}
\label{tab:acc_gc}
\vskip 0.15in
\begin{center}
\resizebox{\textwidth}{!}{%
\begin{tabular}{llcccccccc}
\hline
 &  & \multicolumn{3}{c}{Social Network} && \multicolumn{4}{c}{Bioinformatics} \\ \cline{3-5} \cline{7-10} 
 &  & IMDB-BINARY & IMDB-MULTI & COLLAB && MUTAG & PROTEINS & D\&D & NCI1 \\ \hline
\multicolumn{2}{l}{\# Graphs} & 1,000 & 1,500 & 5,000 && 188 & 1,113 & 1,178 & 4,110 \\
\multicolumn{2}{l}{\# Classes} & 2 & 3 & 3 && 2 & 2 & 2 & 2 \\
\multicolumn{2}{l}{Avg. \# Nodes} & 19.8 & 13.0 & 74.5 && 17.9 & 39.1 & 284.3 & 29.8 \\ \hline
\multirow{2}{*}{Backbones} & GCN & 73.26$\pm$0.46 & 50.39$\pm$0.41 & 80.59$\pm$0.27 && 69.50$\pm$1.78 & 73.24$\pm$0.73 & 72.05$\pm$0.55 & 76.29$\pm$1.79 \\
 & GIN & 72.78$\pm$0.86 & 48.13$\pm$1.36 & 78.19$\pm$0.63 && 81.39$\pm$1.53 & 71.46$\pm$1.66 & 70.79$\pm$1.17 & \textbf{80.00$\pm$1.40} \\ \hline
\multirow{5}{*}{\begin{tabular}[c]{@{}l@{}}Global\\ Pooling\end{tabular}} & Set2Set & 72.90$\pm$0.75 & 50.19$\pm$0.39 & 79.55$\pm$0.39 && 69.89$\pm$1.94 & 73.27$\pm$0.85 & 71.94$\pm$0.56 & 68.55$\pm$1.92 \\
 & SortPool & 72.12$\pm$1.12 & 48.18$\pm$0.83 & 77.87$\pm$0.47 && 71.94$\pm$3.55 & 73.17$\pm$0.88 & 75.58$\pm$0.72 & 73.82$\pm$1.96 \\
 & SAGPool(G) & 72.16$\pm$0.88 & 49.47$\pm$0.56 & 78.85$\pm$0.56 && 76.78$\pm$2.12 & 72.02$\pm$1.08 & 71.54$\pm$0.91 & 74.18$\pm$1.20 \\
 & StructPool & 72.06$\pm$0.64 & 50.23$\pm$0.53 & 77.27$\pm$0.51 && 79.50$\pm$0.75 & 75.16$\pm$0.86 & 78.45$\pm$0.40 & 78.64$\pm$1.53 \\
 & GMT & 73.48$\pm$0.76 & 50.66$\pm$0.82 & 80.74$\pm$0.54 && 83.44$\pm$1.33 & 75.09$\pm$0.59 & \textbf{78.72$\pm$0.59} & 76.35$\pm$2.62 \\ \hline
\multirow{6}{*}{\begin{tabular}[c]{@{}l@{}}Hierarchical\\ Pooling\end{tabular}} & DiffPool & 73.14$\pm$0.70 & 51.31$\pm$0.72 & 78.68$\pm$0.43 && 79.22$\pm$1.02 & 73.03$\pm$1.00 & 77.56$\pm$0.41 & 62.32$\pm$1.90 \\
 & SAGPool(H) & 72.55$\pm$1.28 & 50.23$\pm$0.44 & 78.03$\pm$0.31 && 73.67$\pm$4.28 & 71.56$\pm$1.49 & 74.72$\pm$0.82 & 67.45$\pm$1.11 \\
 & TopKPool & 71.58$\pm$0.95 & 48.59$\pm$0.72 & 77.58$\pm$0.85 && 67.61$\pm$3.36 & 70.48$\pm$1.01 & 73.63$\pm$0.55 & 67.02$\pm$2.25 \\
 & ASAP & 72.81$\pm$0.50 & 50.78$\pm$0.75 & 78.64$\pm$0.5 && 77.83$\pm$1.49 & 73.92$\pm$0.63 & 76.58$\pm$1.04 & 71.48$\pm$0.42 \\
 & MinCutPool & 72.65$\pm$0.75 & 51.04$\pm$0.70 & 80.87$\pm$0.34 && 79.17$\pm$1.64 & 74.72$\pm$0.48 & 78.22$\pm$0.54 & 74.25$\pm$0.86 \\ \cline{2-10} 
 & \textbf{SEP-G} & \textbf{74.12$\pm$0.56} & \textbf{51.53$\pm$0.65} & \textbf{81.28$\pm$0.15} && \textbf{85.56$\pm$1.09} & \textbf{76.42$\pm$0.39} & 77.98$\pm$0.57 & 78.35$\pm$0.33 \\ \hline
\end{tabular}}
\end{center}
\vskip -0.1in
\end{table*}

\subsection{Graph Classification}
Graph classification aims to label a given graph $G$ with the maximum probability among several seed categories. To this end, we need learn the high-level representation from its component nodes, which enables the final classifier evaluate the likelihood of category that the graph belongs to.

\paragraph{Datasets.} Seven benchmarks for graph classification are selected from TU datasets \cite{morris2020tudataset}. Specifically, we employ three social network datasets, including IMDB-BINARY, IMDB-MULTI, and COLLAB; and four bioinformatics datasets, including MUTAG, PROTEINS, D\&D and NCI1. Table~\ref{tab:acc_gc} summarizes the characteristics of the seven employed datasets, and more detailed descriptions can be found in the Appendix A.2.

\paragraph{Baselines.} We first employ two popular backbones in GNNs for comparison, that is, GCN \cite{kipf2017semi} and GIN \cite{xu2019powerful}. Then, we adopt the next five hierarchical pooling methods as baselines: DiffPool \cite{ying2018hierarchical}, SAGPool(H) \cite{lee2019self}, TopKPool \cite{gao2019graph}, ASAP \cite{ranjan2020asap}, and MinCutPool \cite{bianchi2020spectral}. Besides various hierarchical pooling methods, plenty of efforts have also been devoted to global pooling for graph classification. Thus, we also select the following five global pooling techniques for comparison: Set2Set \cite{vinyals15order}, SortPool \cite{zhang2018end}, SAGPool(G) \cite{lee2019self}, StructPool \cite{yuan2020structpool}, and GMT \cite{baek2021accurate}.

\paragraph{Configurations.} Following \cite{xu2019powerful,lee2019self}, 10-fold cross-validation is conducted, and we present the average accuracies achieved to validate the performance of SEP-G in graph classification. In addition, the initial feature inputs is in line with the fair comparison setting in \cite{errica2020fair}. Additional details about experiment setup can be found in the Appendix A.2.

\paragraph{Classification results.} The classification accuracies of SEP-G and other baselines are shown in Table~\ref{tab:acc_gc}, and we can see that our method consistently achieves better or competitive performance as compared to these SOTA methods. In particular, SEP-G obtains a unified improvement in social network datasets, which differs from the performance in bioinformatics datasets. This performance divergence may be because SEP only relies on the network structure for hierarchical pooling, while the structural information in social network datasets is more redundant than that in bioinformatics datasets \cite{centola2010spread}.
It is worth noting that there are not any pooling methods suppress GIN in NCI1, or, put differently, pooling methods also do not show unified promotion in comparison with backbones. Considering the recent finding that message-passing is the crucial mechanism in graph classification \cite{mesquita2020rethinking}, this phenomenon may not be so disappointing.

\begin{table*}[!ht]
\centering
\caption{\textbf{Graph classification accuracies of SEP-G with various backbones.} The default backbone is GCN, and we denote it as SEP-G-GCN for a better illustration.}
\label{tab:sep-variants}
\vskip 0.15in
\resizebox{\textwidth}{!}{%
\begin{tabular}{lcccccccc}
\hline
\multirow{2}{*}{Variants} & \multicolumn{3}{c}{Social Network} &  & \multicolumn{4}{c}{Bioinformatics} \\ \cline{2-4} \cline{6-9} 
 & IMDB-BINARY & IMDB-MULTI & COLLAB &  & MUTAG & PROTEINS & DD & NCI1 \\ \hline
SEP-G-GCN & \textbf{74.12$\pm$0.56} & 51.53$\pm$0.65 & \textbf{81.28$\pm$0.15} &  & \textbf{85.56$\pm$1.09} & 76.42$\pm$0.39 & 77.98$\pm$0.57 & 78.35$\pm$0.33 \\
SEP-G-GIN & 73.37$\pm$0.95 & 51.81$\pm$0.98 & 79.18$\pm$0.60 &  & 83.22$\pm$1.28 & 74.77$\pm$1.42 & 75.98$\pm$1.15 & 76.59$\pm$1.65 \\
SEP-G-GAT & 73.24$\pm$0.81 & \textbf{51.87$\pm$0.45} & 79.26$\pm$0.39 &  & 84.45$\pm$1.81 & \textbf{76.72$\pm$0.92} & \textbf{78.07$\pm$0.74} & \textbf{78.43$\pm$1.07} \\
SEP-G-ChebNet & 73.72$\pm$0.42 & 50.84$\pm$0.68 & 80.73$\pm$0.43 &  & 83.25$\pm$1.13 & 74.67$\pm$0.75 & 76.69$\pm$0.71 & 77.68$\pm$0.97 \\
SEP-G-GraphSAGE & 73.14$\pm$0.87 & 50.43$\pm$1.31 & 79.88$\pm$0.58 &  & 83.75$\pm$1.43 & 75.26$\pm$0.86 & 77.95$\pm$0.55 & 77.65$\pm$1.21 \\ \hline
\end{tabular}%
}
\vskip -0.1in
\end{table*}

\paragraph{Variants of SEP-G.} As discussed in Section~\ref{sec:preliminaries}, besides GCN, our proposed pooling operator can also work with other GNNs like GAT and GIN. Here, we delve deeper into the collaboration ability of our pooling method with other GNNs. Specifically, we employ the ChebNet \cite{defferrard2016convolutional}, GraphSAGE \cite{hamilton2017inductive}, GAT \cite{velivckovic2018graph} and GIN \cite{xu2019powerful}. The classification results are shown in Table~\ref{tab:sep-variants}. As can be seen, the overall superior performance is obtained by these variants, which suggests the effectiveness and kindness of SEP. Notably, a better performance on IMDB-MULTI, PROTEINS, DD and NCI1 is obtained by SEP with GAT, which further indicates the huge potential of SEP in collaboration with other SOTA backbones.

\begin{figure}[!h]
  \vskip 0.2in
  \centering
    \begin{subfigure}{.15\textwidth}
    \centering
    \includegraphics[width=\linewidth]{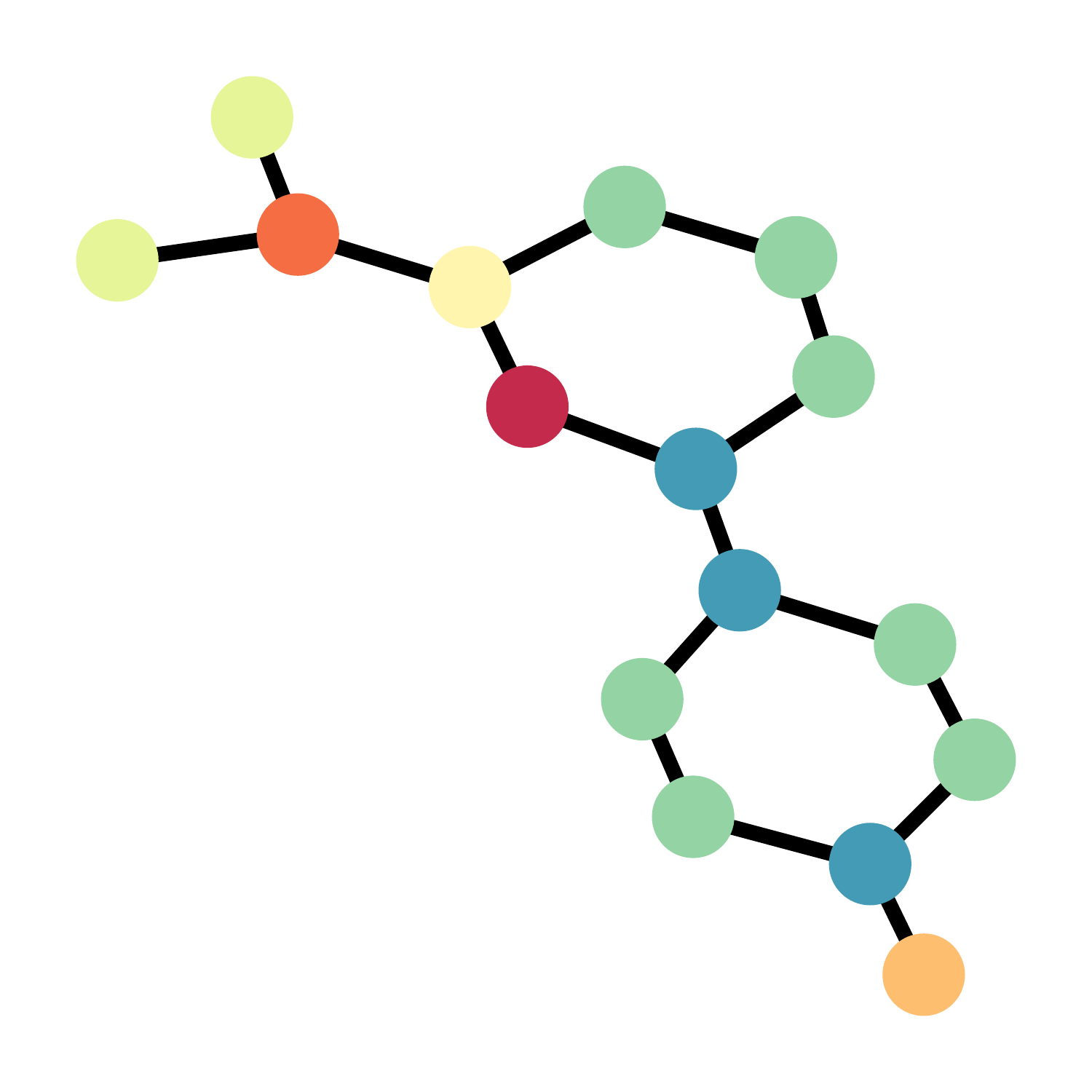}
    % \caption{DiffPool}
  \end{subfigure}
  \begin{subfigure}{.15\textwidth}
    \centering
    \includegraphics[width=\linewidth]{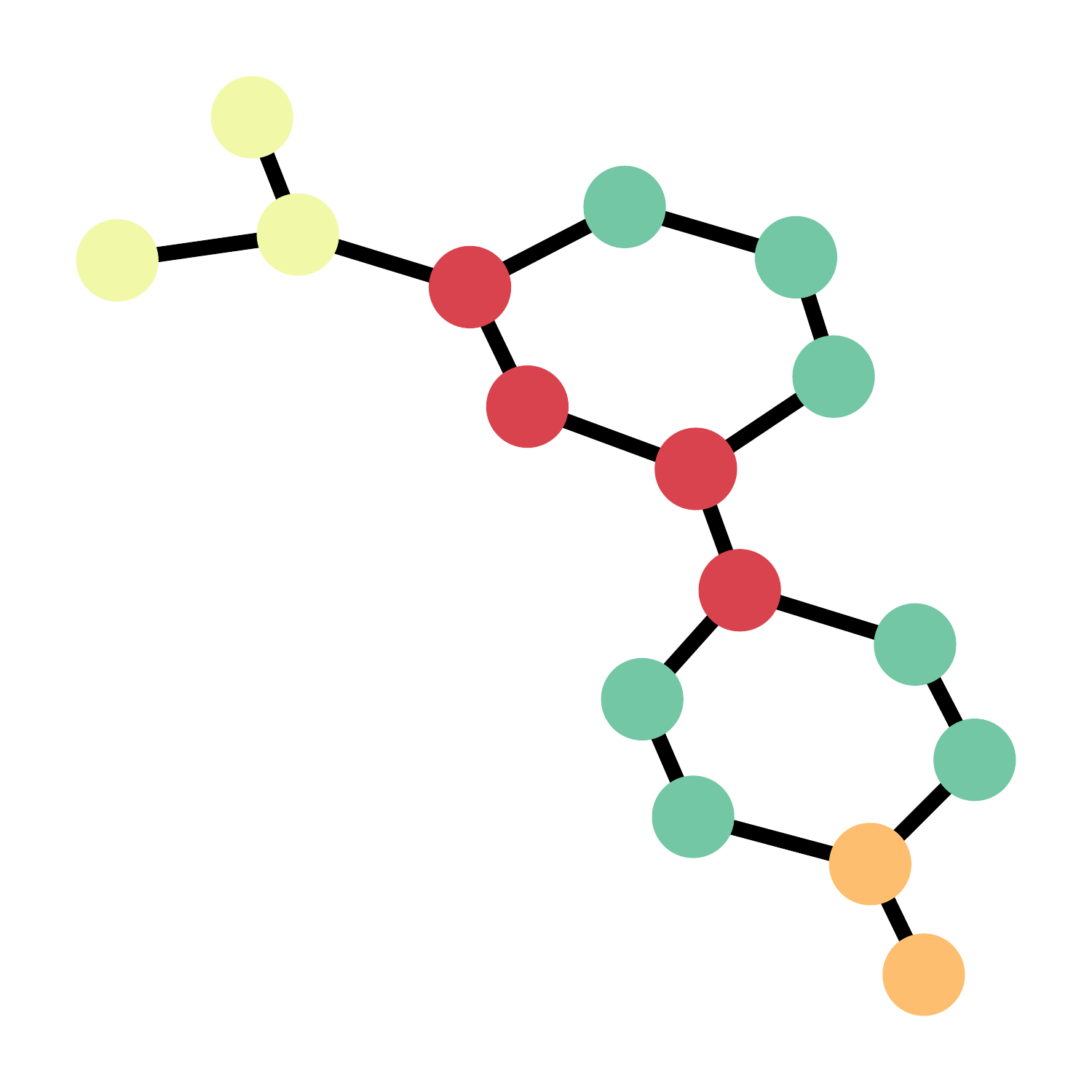}
    % \caption{MinCutPool}
  \end{subfigure} 
  \begin{subfigure}{.15\textwidth}
    \centering
    \includegraphics[width=\linewidth]{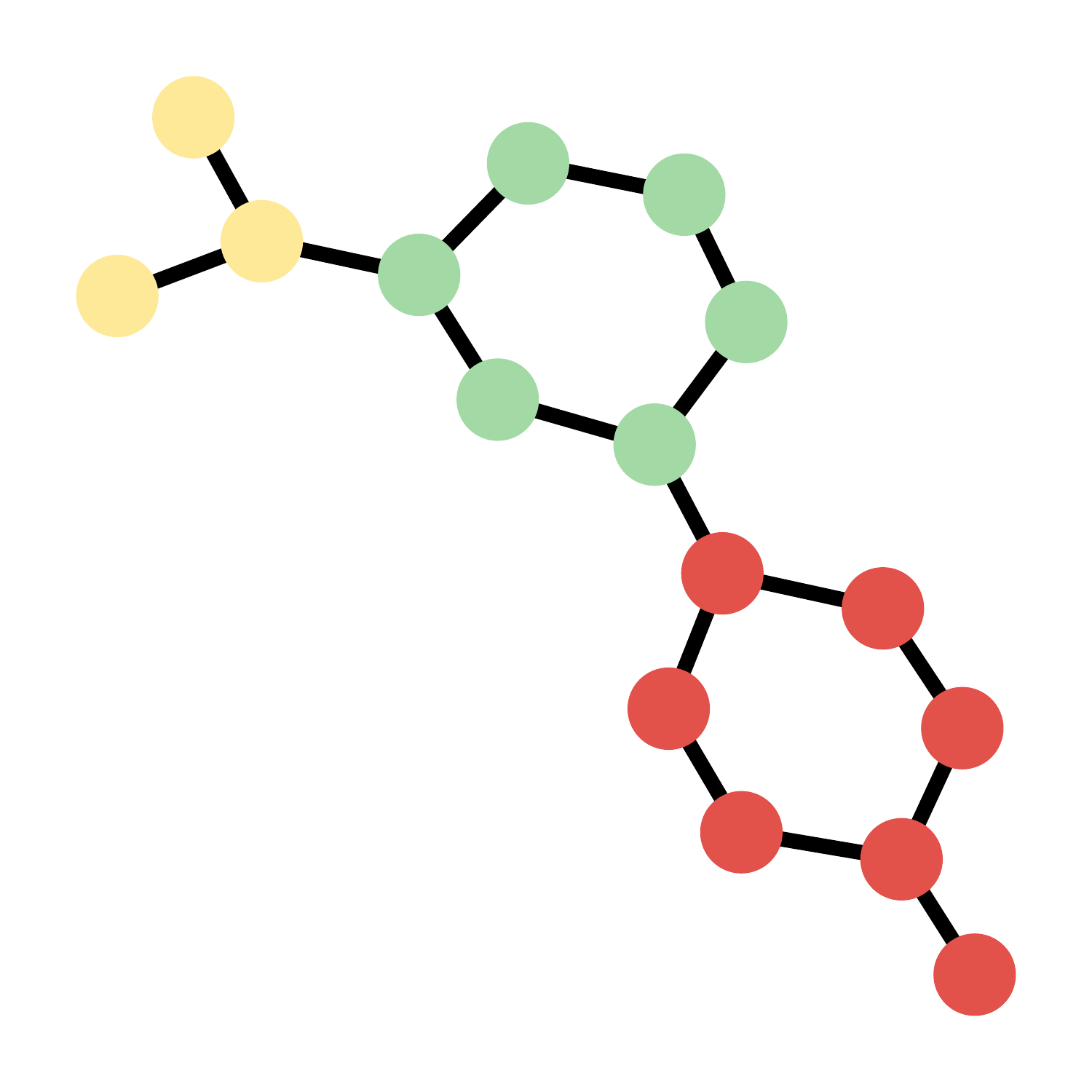}
    % \caption{SEP}
  \end{subfigure} 
  \\
  \begin{subfigure}{.15\textwidth}
    \centering
    \includegraphics[width=0.9\linewidth]{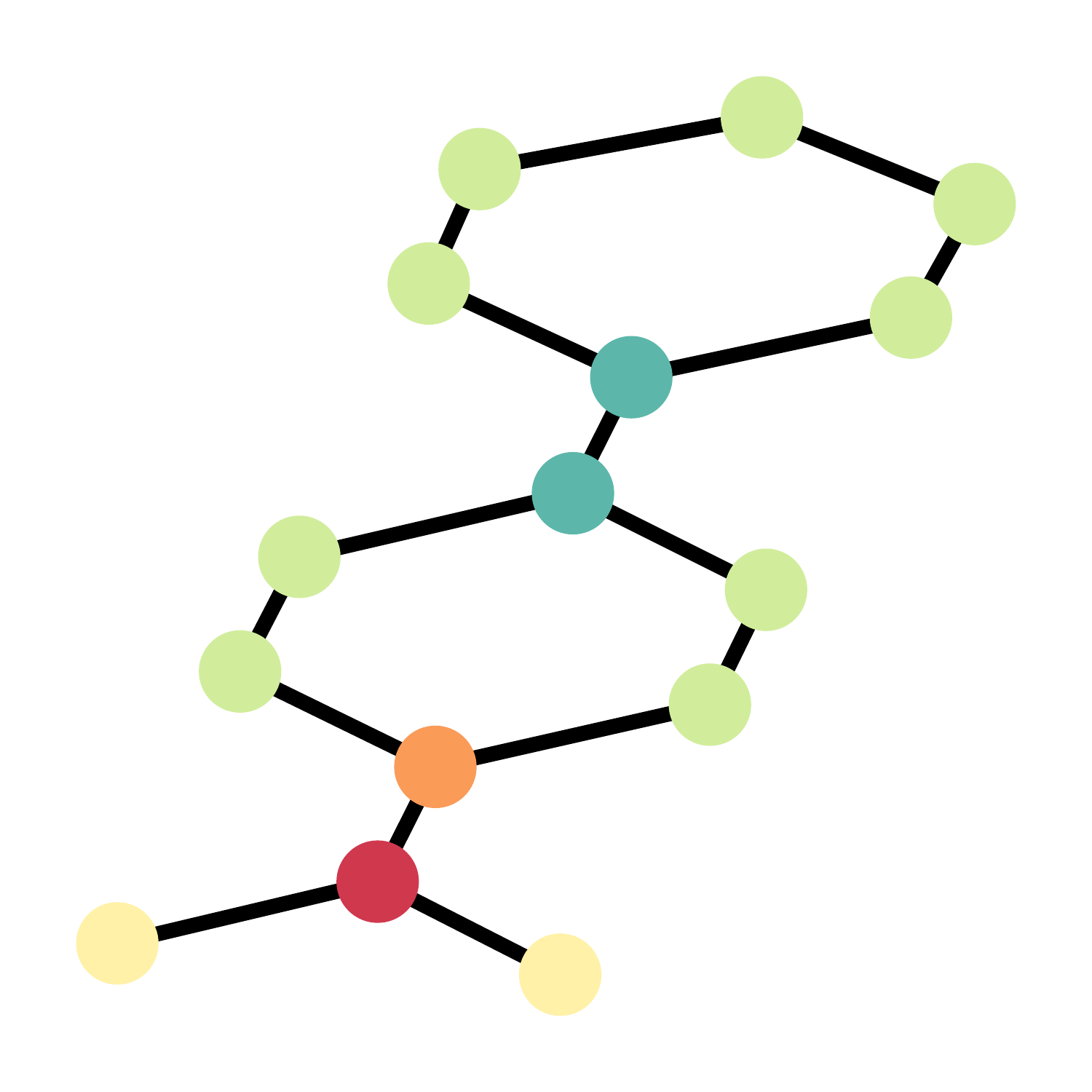}
    \vspace{5pt}
    \caption{DiffPool}
  \end{subfigure}
  \begin{subfigure}{.15\textwidth}
    \centering
    \includegraphics[width=0.9\linewidth]{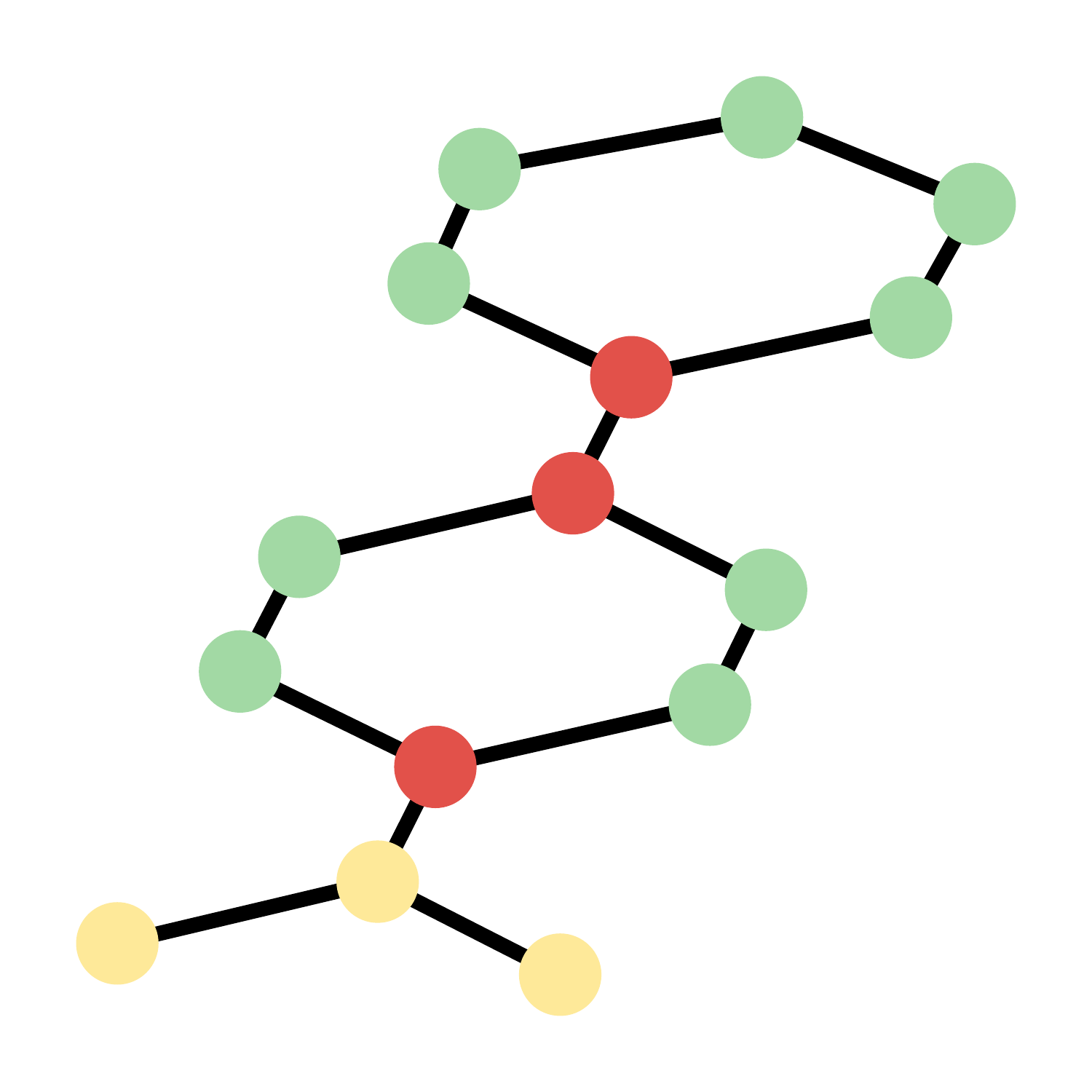}
    \vspace{5pt}
    \caption{MinCutPool}
  \end{subfigure} 
  \begin{subfigure}{.15\textwidth}
    \centering
    \includegraphics[width=0.9\linewidth]{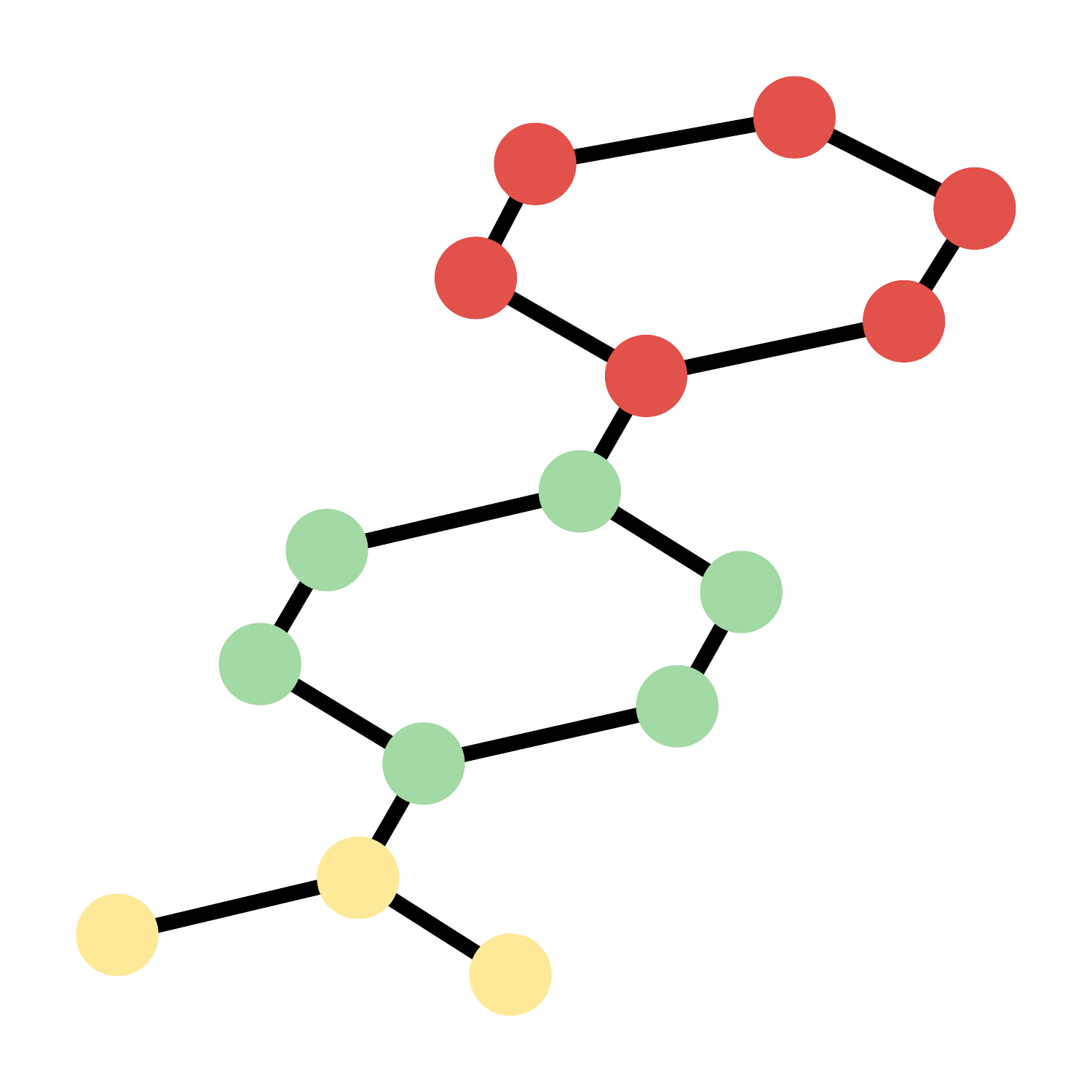}
    \vspace{5pt}
    \caption{SEP}
  \end{subfigure} 
  \caption{Essential structure preserving on MUTAG.} 
  \label{fig:visual_case}
  \vskip 0.2in
\end{figure}

\paragraph{Visualization case study.} To better demonstrate the effectiveness of SEP on essential structure preserving, we present the clustering results of DiffPool, MinCutPool, and SEP on samples from the MUTAG dataset after the first graph coarsening. As shown in Figure \ref{fig:visual_case}, DiffPool and MinCutPool severely damage the essential structures of the two compounds, in which the ring structures of two molecular formulas are arbitrarily torn into several pieces.
Fortunately, SEP manages to take good care of these vital structures in the process of cluster assignment generation, and this shows the effectiveness of SEP in issues addressing.

\subsection{Node Classification}
Node classification is another important task regarding graphs in GNNs, which aims to label each node in a given graph $G$. Here, we conduct corresponding experiments on ubiquitous benchmarks to validate the effectiveness of our proposed SEP-N.

% Attributed Graph Clustering via Adaptive Graph Convolution
\paragraph{Datasets.} We evaluate SEP-N under the transductive learning setting, which includes three datasets Cora, Citeseer and Pubmed \cite{sen2008collective}. The three benchmarks are constructed on the connections of document citations, which means nodes are documents and edges are citation links. The node features are different among three datasets. Specifically, the input features of Cora and Citeseer are one-hot embedding of words in each document, while the node features of Pubmed are the TF-IDF weighted word vectors.
Following the experiment setup in previous works \cite{kipf2017semi,gao2019graph}, the designated training/validation/testing splits on Cora, Citeseer and Pubmed are adopted, that is, training set has 20 nodes for each class, validation set has 500 nodes and testing set has 1,000 nodes. Table~\ref{tab:acc-nc} shows the dataset statistics.

\paragraph{Baselines.} In addition to g-U-nets \cite{gao2019graph} that has the same encoder-decoder design, we also include three SOTA backbones in GNNs: GCN \cite{kipf2017semi}, GAT \cite{velivckovic2018graph}, and GIN \cite{xu2019powerful}. Moreover, other following works based on the three models are also employed for a comprehensive comparison, such as FastGCN \cite{chen2018fastgcn}, APPNP \cite{klicpera2019predict}, MixHop \cite{abu2019mixhop}, SGC \cite{wu2019simplifying}, DGI \cite{velickovic2019deep}, S$^2$GC \cite{zhu2020simple}, and GCNII \cite{chen2020simple}.

\paragraph{Configurations.} For node classification tasks, we fix our SEP-N with two blocks of encoders and decoders as presented in Figure~\ref{fig:sep-unet}, and plan to obtain the coding tree for each dataset under the guidance of three-dimensional structural entropy. In particular, each block has a GCN layer followed by a SEP (or SEP-U) layer. Finally, a GCN layer is adopted to make final prediction. There is no doubt that skip connections between layers are also established between encoder and decoder analogous to g-U-nets. Note that we add an additional linear layer after each SEP or SEP-U layer to learn more task-specific node representations. 
Dropout \cite{srivastava2014dropout} with ReLU on feature matrices is applied to all layers in SEP-N, and L2 regularization is also adopted to avoid over-fitting. Detailed descriptions for experimental setup are shown in Appendix A.3.

\paragraph{Classification results.} The accuracies of our proposed SEP-N and baselines on three benchmarks are shown in Table~\ref{tab:acc-nc}. In general, we can observe that the SEP-N does not achieve the best performance on any datasets, which are shown in S$^2$GC on Citeseer and GCNII on Cora and Pubmed. 
This problem may be attributed to the cluster assignments generation, which only relies on structural information, while the tasks regarding node classification are more dependent on the input features of nodes.
However, SEP-N still obtains competitive results with only 5 GCN layers, which is much less than the design of S$^2$GC (16) and GCNII (64 for Cora, 32 for Citeseer, 16 for Pubmed).
In particular, the accuracies of SEP-N on three datasets are consistently better than the three backbones (i.e., GCN, GAT and GIN). Furthermore, compared with g-U-nets, which is also designed with GCN and hierarchical pooling mechanism, our method also surpasses it on two out of three datasets (Cora and Pubmed) even with fewer learning blocks. Note that superior performance of SEP-N on Citeseer has also been achieved with different numbers of blocks, which we will describe it in the next experiments. In summary, we can confirm the effectiveness of the proposed hierarchical pooling operation in node representation learning.

\begin{table}[!ht]
\caption{\textbf{Node classification accuracies on Cora, Citeseer, and Pubmed (\%).} We highlight our results and those that are significantly higher than all other methods.}
\label{tab:acc-nc}
\vskip 0.15in
\begin{center}
\begin{tabular}{lccc}
\hline
 & Cora & Citeseer & Pubmed \\ \hline
\# Nodes & 2,708 & 3,327 & 19,717 \\
\# Edges & 5,429 & 4,732 & 44,338 \\
\# Features & 1,433 & 3,703 & 4,500 \\
\# Classes & 7 & 6 & 3 \\ \hline
GCN & 81.4$\pm$0.4 & 70.9$\pm$0.5 & 79.0$\pm$0.4 \\
GAT & 83.3$\pm$0.7 & 72.6$\pm$0.6 & 78.5$\pm$0.3 \\
GIN & 77.6$\pm$1.1 & 66.1$\pm$0.9 & 77.0$\pm$1.2 \\
FastGCN & 79.8$\pm$0.3 & 68.8$\pm$0.6 & 77.4$\pm$0.3 \\
APPNP & 83.3$\pm$0.5 & 71.7$\pm$0.6 & 80.1$\pm$0.2 \\
MixHop & 81.8$\pm$0.6 & 71.4$\pm$0.8 & 80.0$\pm$1.1 \\
DGI & 82.5$\pm$0.7 & 71.6$\pm$0.7 & 78.4$\pm$0.7 \\
SGC & 81.0$\pm$0.03 & 71.9$\pm$0.11 & 78.9$\pm$0.01 \\
S$^2$GC & 83.5$\pm$0.02 & \textbf{73.6$\pm$0.09} & 80.2$\pm$0.02 \\
GCNII & \textbf{85.5$\pm$0.5} & 73.4$\pm$0.6 & \textbf{80.3}$\pm$0.4 \\
g-U-Nets & 84.4$\pm$0.6 & 73.2$\pm$0.5 & 79.6$\pm$0.2 \\ \hline
\textbf{SEP-N} & \textbf{84.8$\pm$0.4} & \textbf{72.9$\pm$0.7} & \textbf{80.2$\pm$0.8} \\ \hline
\end{tabular}%
\end{center}
\vskip -0.1in
\end{table}

\begin{table*}[!tbp]
\caption{\textbf{Node classification accuracies with different network depths (\%).} We highlight the best results with different depths for SEP-N and g-U-Nets.}
\label{tab:com-nc-depth}
\vskip 0.15in
\begin{center}
\begin{tabular}{lcc|cc|cc}
\hline
 & \multicolumn{2}{c|}{Cora} & \multicolumn{2}{c|}{Citeseer} & \multicolumn{2}{c}{Pubmed} \\ \hline
Depth & g-U-Nets & SEP-N & g-U-Nets & SEP-N & g-U-Nets & SEP-N \\ \hline
1 & $-$ & 84.3$\pm$0.6& $-$ & \textbf{73.3$\pm$0.6} & $-$ & 78.9$\pm$0.6 \\
2 & 82.6$\pm$0.6 & \textbf{84.8$\pm$0.4} & 71.8$\pm$0.5 & 72.9$\pm$0.7 & 79.1$\pm$0.3 & \textbf{80.2$\pm$0.8} \\
3 & 83.8$\pm$0.7 & 84.5$\pm$0.3 & 72.7$\pm$0.7 & 72.1$\pm$0.6 & 79.4$\pm$0.4 & 79.5$\pm$0.5 \\
4 & \textbf{84.4$\pm$0.6} & 83.6$\pm$0.6 & \textbf{73.2$\pm$0.5} & 72.1$\pm$0.2 & \textbf{79.6$\pm$0.2} & 78.5$\pm$0.3 \\
5 & 84.1$\pm$0.5 & 83.9$\pm$0.5 & 72.8$\pm$0.6 & 72.4$\pm$0.6 & 79.5$\pm$0.3 & 79.8$\pm$0.7 \\ \hline
\end{tabular}%
\end{center}
\vskip -0.1in
\end{table*}

\paragraph{Ablation Study of Network Depth.} Besides the other bunch of hyper-parameters in GNNs, the network depth, corresponding to the number of blocks, is also another crucial setting in model construction. We, thus, delve deeper into the effect of model depth on node classification performance. We iteratively test SEP-N with various numbers of blocks $\in\{1, 2, 3, 4, 5\}$, and the results are shown in Table~\ref{tab:com-nc-depth}.
As we mentioned above, SEP-N achieves superior performance on Citeseer with only one encoder and decoder.
In particular, compared with g-U-Nets, we can see that the best performance of SEP-N on three datasets is achieved with at most 2 blocks, which once again proves the capacity of shallow networks in high-level feature encoding \cite{gao2019graph}. Moreover, this scene also reveals that additional connection information among layers for cluster assignments generation is helpful for representation learning, as well as benefits model optimization that better performance with fewer parameters and computations.

\section{Conclusions}
In this paper, we develop an optimization algorithm to address several limitations of existing hierarchical pooling approaches. In particular, under the guidance of structural entropy minimization, our pooling method, SEP, can not only capture the connectivities among pooling layers but also fix the problem of destroying local structure due to the hyper-parameter for node compression.
Based on the proposed SEP, we introduce two learning models, SEP-G and SEP-N, for graph classification and node classification, respectively.
Experimental results suggest that SEP-G achieves significant improvements on graph classification, and SEP-N obtains superior performance as compared to other GNNs on node classification tasks.
An interesting direction for future work is shown in the combination of structural entropy and node features.

\section*{Acknowledgements}
This research was supported by NSFC (Grant No. 61932002).

% \textbf{Do not} include acknowledgements in the initial version of
% the paper submitted for blind review.

% If a paper is accepted, the final camera-ready version can (and
% probably should) include acknowledgements. In this case, please
% place such acknowledgements in an unnumbered section at the
% end of the paper. Typically, this will include thanks to reviewers
% who gave useful comments, to colleagues who contributed to the ideas,
% and to funding agencies and corporate sponsors that provided financial

% \newpage
\bibliography{example_paper}
\bibliographystyle{icml2022}

%%%%%%%%%%%%%%%%%%%%%%%%%%%%%%%%%%%%%%%%%%%%%%%%%%%%%%%%%%%%%%%%%%%%%%%%%%%%%%%
%%%%%%%%%%%%%%%%%%%%%%%%%%%%%%%%%%%%%%%%%%%%%%%%%%%%%%%%%%%%%%%%%%%%%%%%%%%%%%%
% APPENDIX
%%%%%%%%%%%%%%%%%%%%%%%%%%%%%%%%%%%%%%%%%%%%%%%%%%%%%%%%%%%%%%%%%%%%%%%%%%%%%%%
%%%%%%%%%%%%%%%%%%%%%%%%%%%%%%%%%%%%%%%%%%%%%%%%%%%%%%%%%%%%%%%%%%%%%%%%%%%%%%%
\newpage
\appendix
\onecolumn

\setcounter{table}{0}
\setcounter{figure}{0}
\renewcommand{\thetable}{A.\arabic{table}}
\renewcommand{\thefigure}{A.\arabic{figure}}

\section{Experiment Setup}
In this section, we introduce the experimental details about graph reconstruction, graph classification, and node classification tasks respectively.

\subsection{Graph Reconstruction}
\paragraph{Dataset.} Graph reconstruction experiments with synthetic graphs represented in a 2-D Euclidean space, such as ring and grid structures. The node features of a graph consist of their location in a 2-D coordinate space, and the adjacency matrix indicates the connectivity pattern of nodes. The goal here is to restore all node locations from compressed features after pooling, with the intact adjacency matrix.

\begin{figure}[!ht]
\vskip 0.2in
\begin{center}
\centerline{\includegraphics[width=0.5\columnwidth]{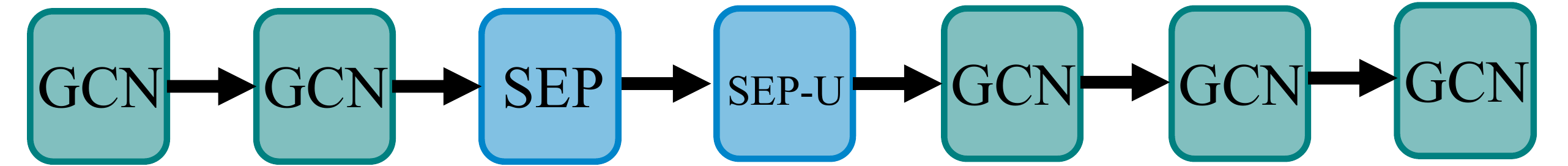}}
\caption{The architecture for graph reconstruction.} 
\label{fig:sep-gcn-r}
\end{center}
\vskip -0.2in
\end{figure}

\paragraph{Implementation details.} Following \cite{bianchi2020spectral}, we use the two message passing layers both right before the pooling operation and right after the unpooling operation. Also, both pooling and unpooling operations are performed once and sequentially connected, as illustrated in the Figure \ref{fig:sep-gcn-r}. 
We compare our methods against both the node drop methods, including TopKPool \cite{gao2019graph}, SAGPool \cite{lee2019self} and ASAP \cite{ranjan2020asap}, and node clustering methods, including DiffPool \cite{ying2018hierarchical} and MinCutPool \cite{bianchi2020spectral}. 
For the node drop methods, we use the unpooling operation proposed in the graph U-net \cite{gao2019graph}. For the node clustering methods, we use the graph coarsening schemes described in the Equation 15, with their specific implementations on generating an assignment matrix. 
For our proposed method, we follow the setting of node clustering methods. In particular, we finely tuned the height of coding tree produced by Algorithm 1 to make sure the number of nodes compressed in the first layer close to the setting of baselines.
For model configuration, the pooling ratio of all models is set to 25\%, the learning rate is set to $5\times 10^{-3}$, and hidden size is set to 32. For the loss function, we use the Mean Squared Error (MSE) to train models. We then optimize the network with Adam optimizer. We use the early stopping criterion, where we stop the training if there is no further improvement on the training loss during 1,000 epochs. Further, the maximum number of epochs is set to 10,000. Note that, there is no other available graphs for validation of the synthetic graph, such that we train and test the models only with the given graph in the Figure 3(a).

\begin{figure}[!ht]
  \vskip 0.2in
  \begin{center}
  \begin{subfigure}{.25\textwidth}
    \centering
    \includegraphics[width=\linewidth]{topk-ring.pdf}
    \caption{TopKPool}
  \end{subfigure} 
  \begin{subfigure}{.25\textwidth}
    \centering
    \includegraphics[width=\linewidth]{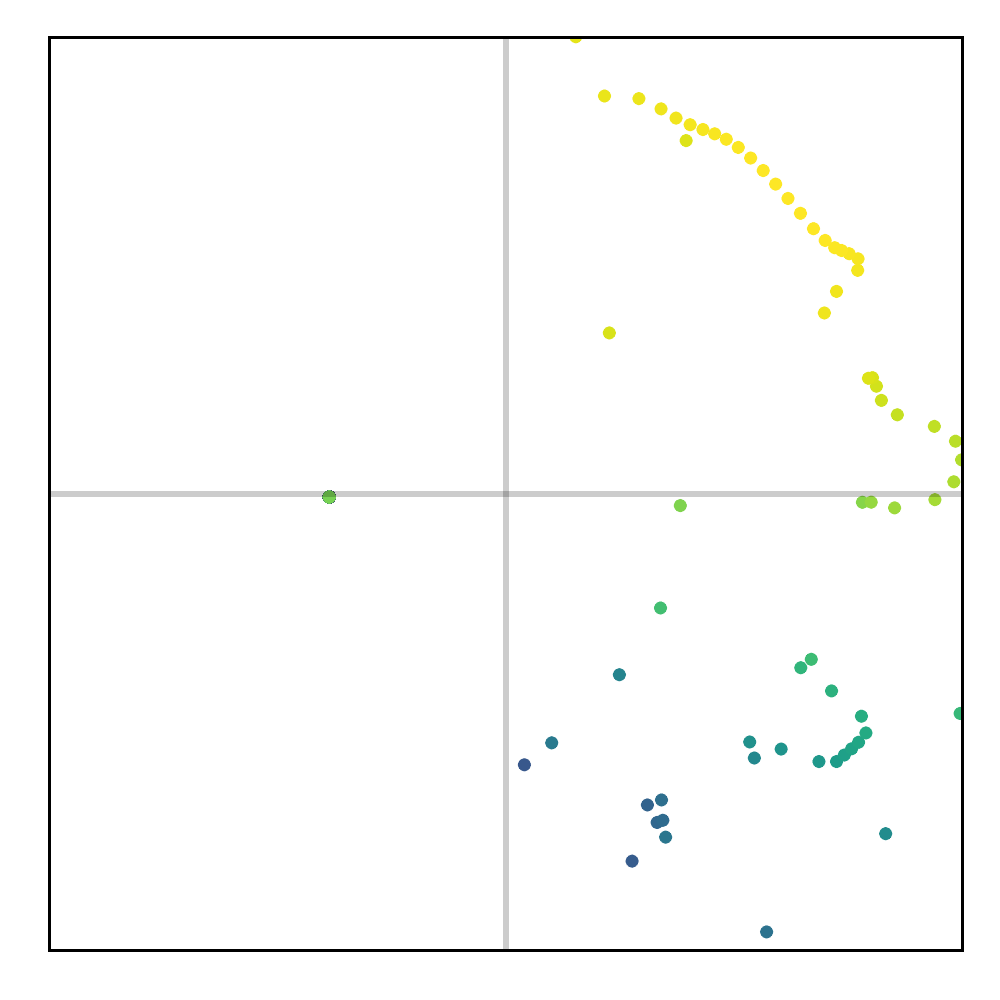}
    \caption{SAGPool}
  \end{subfigure} 
  \begin{subfigure}{.25\textwidth}
    \centering
    \includegraphics[width=\linewidth]{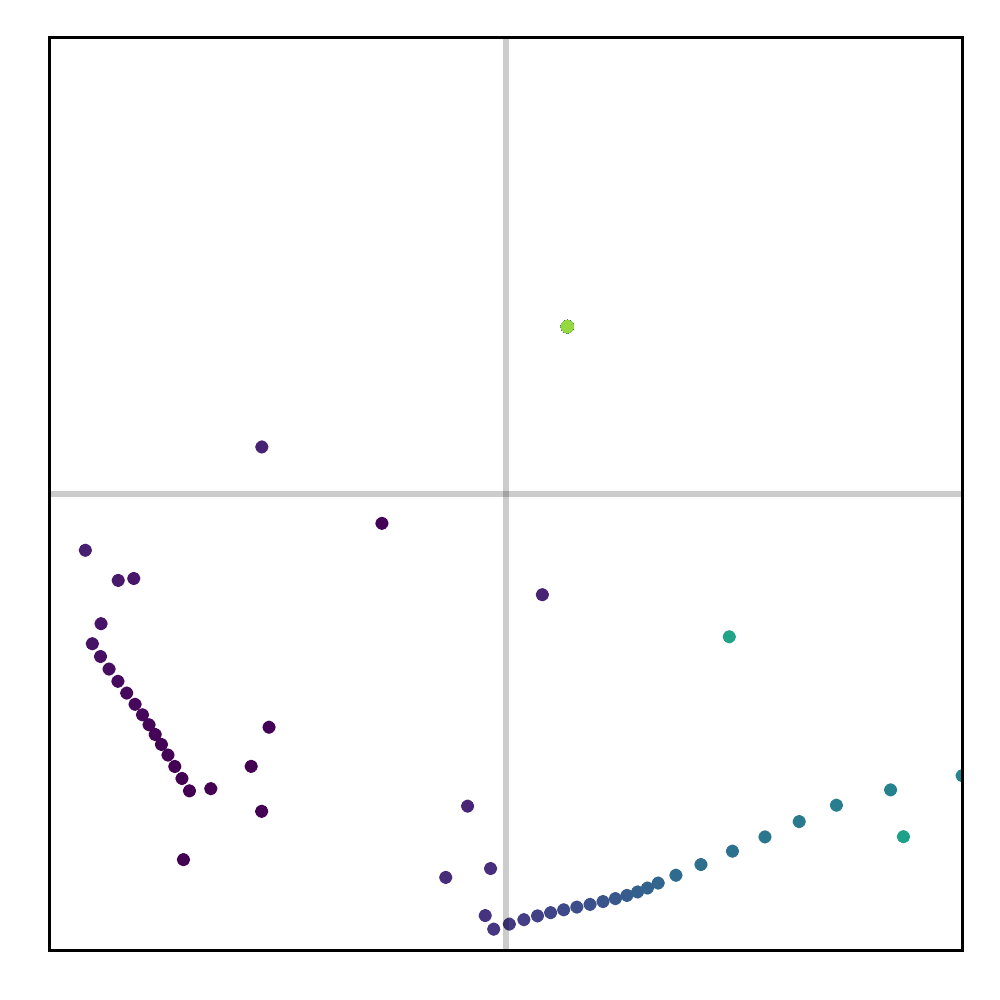}
    \caption{ASAPPool}
  \end{subfigure} \\
  \begin{subfigure}{.255\textwidth}
    \centering
    \includegraphics[width=\linewidth]{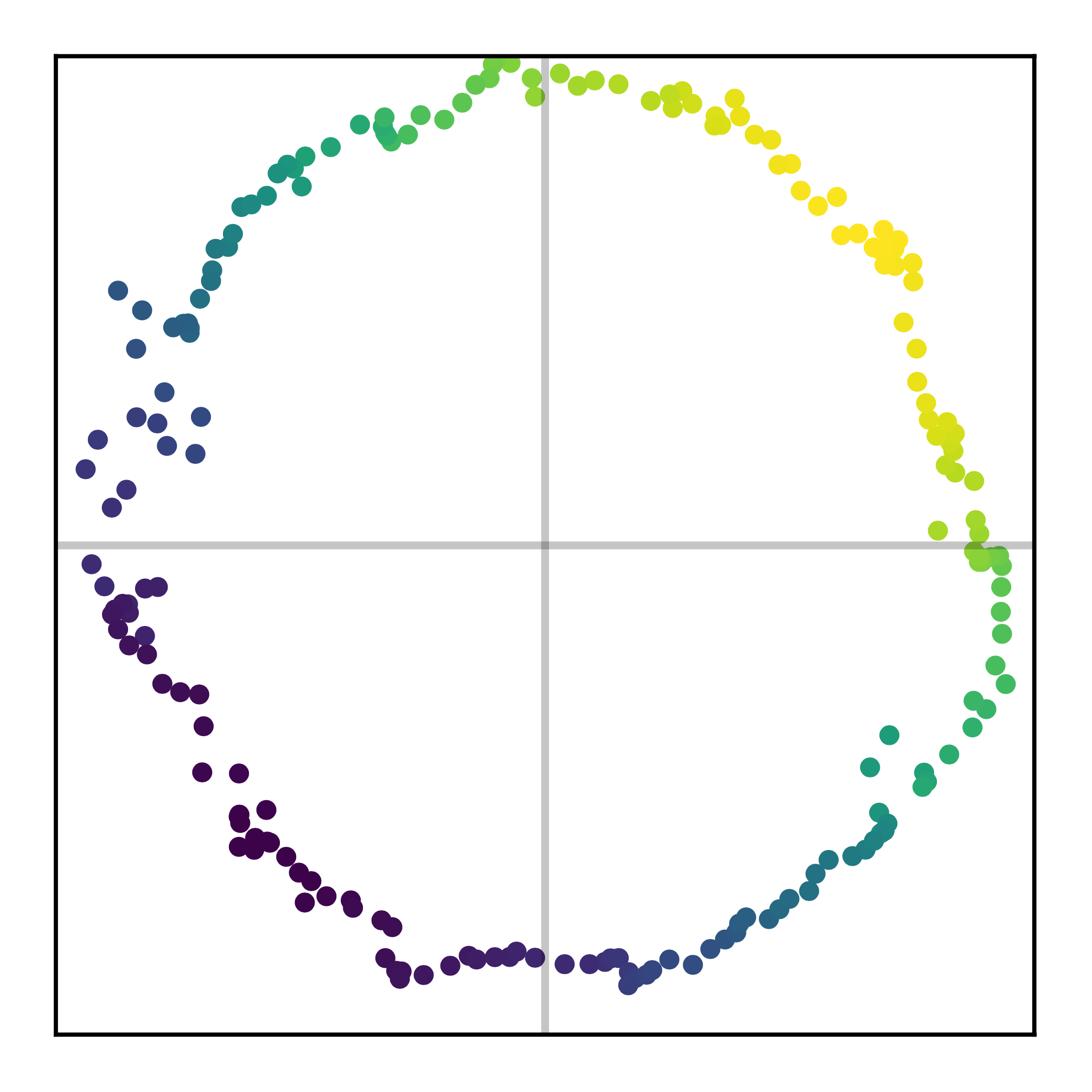}
    \caption{DiffPool}
  \end{subfigure} 
  \begin{subfigure}{.25\textwidth}
    \centering
    \includegraphics[width=\linewidth]{minCutPool-ring.pdf}
    \caption{minCutPool}
  \end{subfigure} 
  \begin{subfigure}{.25\textwidth}
    \centering
    \includegraphics[width=\linewidth]{SEP-U-ring.pdf}
    \caption{SEP}
  \end{subfigure}
  \caption{Reconstruction results of ring synthetic graphs, compared to node drop and clustering methods.} 
  \label{fig:resconstruction-full-ring}
  \end{center}
  \vskip -0.2in
\end{figure}

\begin{figure}[!ht]
  \vskip 0.2in
  \begin{center}
  \begin{subfigure}{.25\textwidth}
    \centering
    \includegraphics[width=\linewidth]{topk-grid.pdf}
    \caption{TopKPool}
  \end{subfigure} 
  \begin{subfigure}{.25\textwidth}
    \centering
    \includegraphics[width=\linewidth]{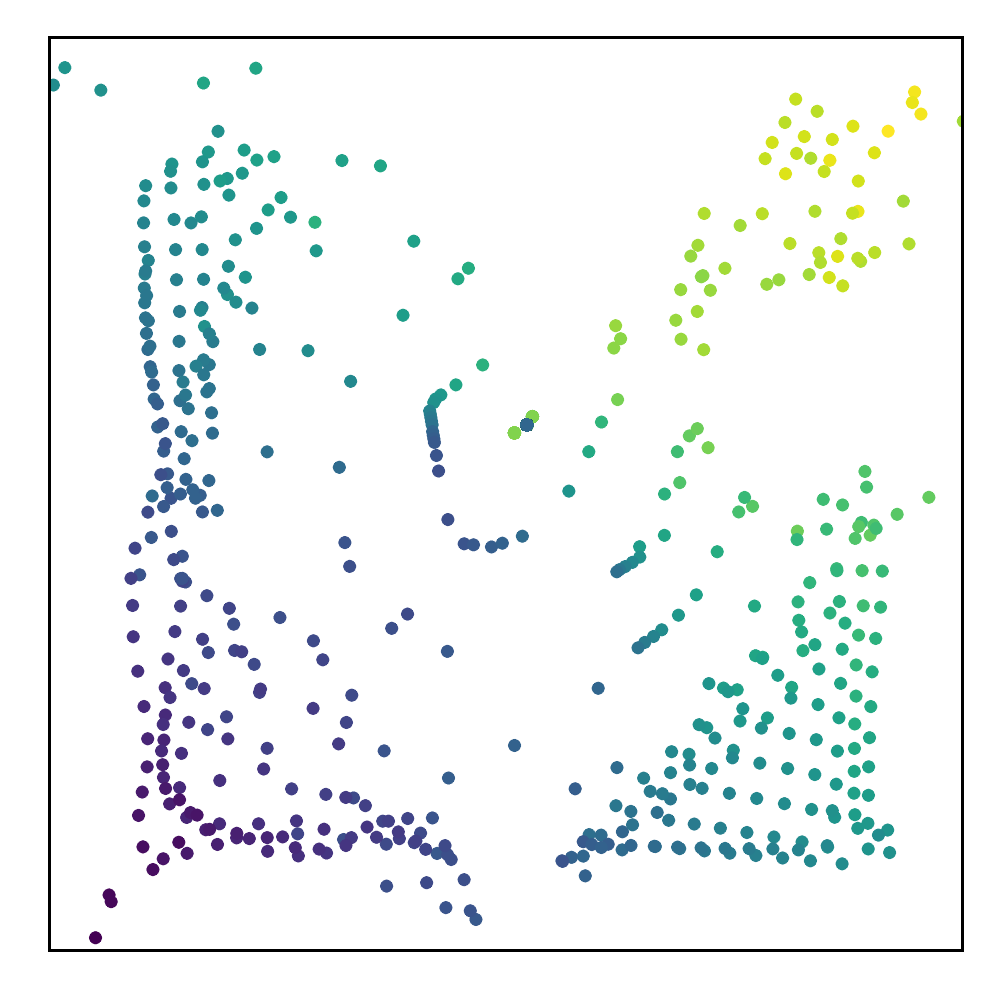}
    \caption{SAGPool}
  \end{subfigure} 
  \begin{subfigure}{.25\textwidth}
    \centering
    \includegraphics[width=\linewidth]{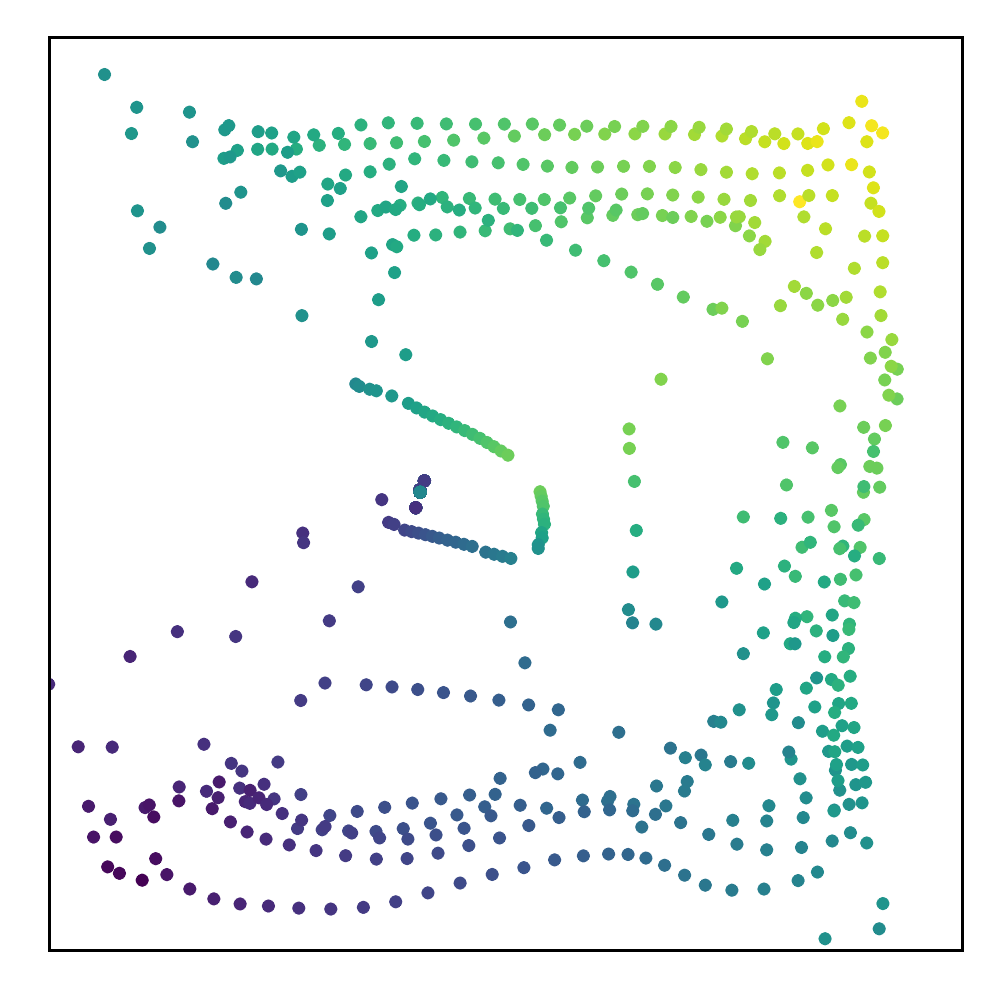}
    \caption{ASAPPool}
  \end{subfigure} \\
  \begin{subfigure}{.255\textwidth}
    \centering
    \includegraphics[width=\linewidth]{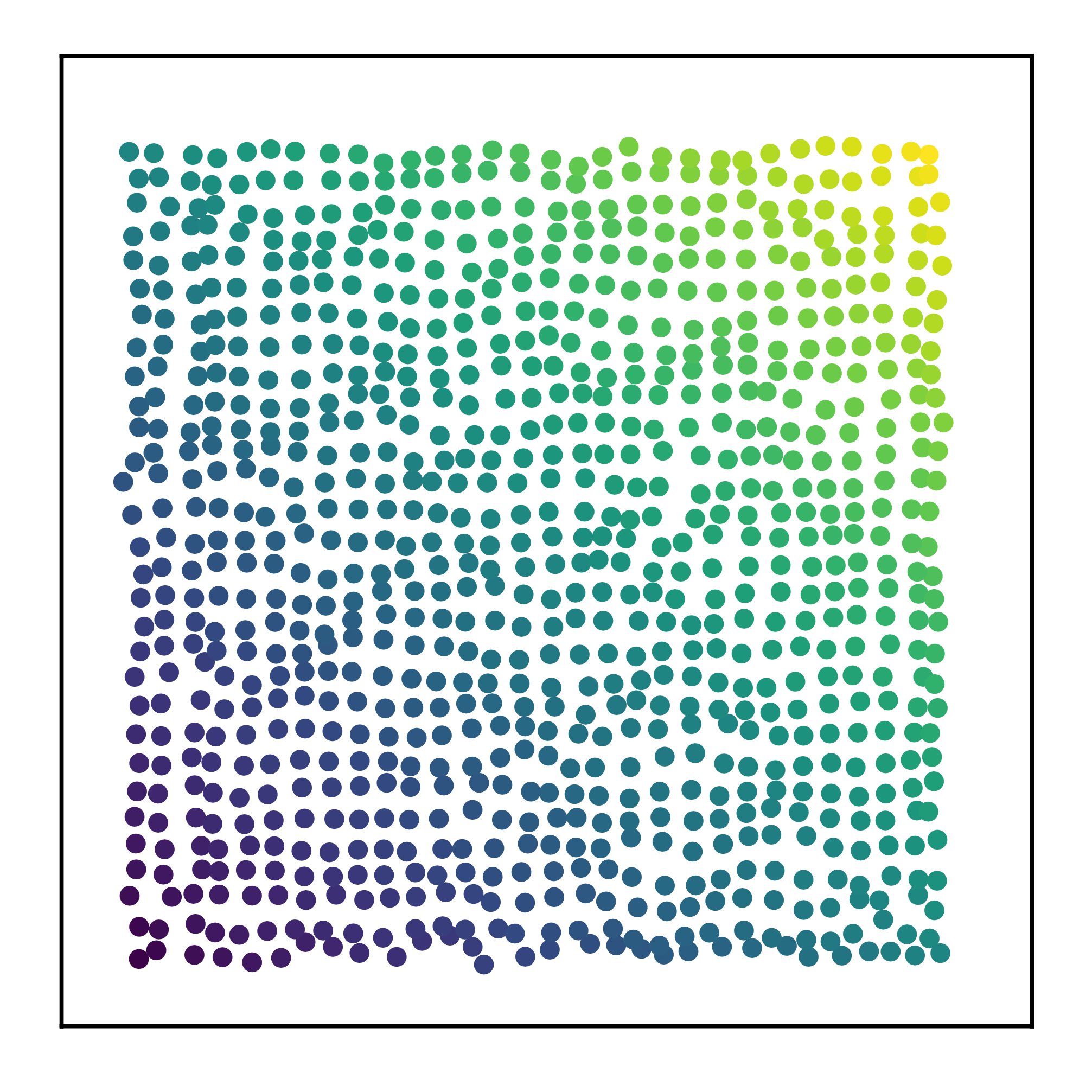}
    \caption{DiffPool}
  \end{subfigure} 
  \begin{subfigure}{.25\textwidth}
    \centering
    \includegraphics[width=\linewidth]{minCutPool-grid.pdf}
    \caption{minCutPool}
  \end{subfigure} 
  \begin{subfigure}{.25\textwidth}
    \centering
    \includegraphics[width=\linewidth]{SEP-U-grid.pdf}
    \caption{SEP}
  \end{subfigure}
  \caption{Reconstruction results of grid synthetic graphs, compared to node drop and clustering methods.} 
  \label{fig:resconstruction-full-grid}
  \end{center}
  \vskip -0.2in
\end{figure}

\paragraph{Reconstruction results on all hierarchical pooling methods.} Figure~\ref{fig:resconstruction-full-ring} and \ref{fig:resconstruction-full-grid} show the results of reconstructed graphs by model with various pooling methods. 
As can be seen from the first row of Figure~\ref{fig:resconstruction-full-ring} and \ref{fig:resconstruction-full-grid}, the node drop methods suffer from the issue of information loss, resulting in the basic shape of original graphs can not be identified. 
For node clustering methods, DiffPool and MinCutPool, the basic shape of original graphs is basically retained.
However, we can still see the significant distortion in the edge of ring and the center of grid, which are almost prevented in SEP. 

\subsection{Graph Classification}
\paragraph{Social network datasets.} IMDB-BINARY and IMDB-MULTI are derived from the collaboration of a movie set. In these two datasets, every graph consists of actors or actresses, and each edge between two nodes represents their cooperation in a certain movie. Each graph is derived from a prespecified movie, and its label corresponds to the genre of this movie. Similarly, COLLAB is also a collaboration dataset but from a scientific realm, which includes three public collaboration datasets (i.e., Astro Physics, High Energy Physics and Condensed Matter Physics). Many researchers from each field form various ego networks for the graphs in this benchmark. The label of each graph is the research field to which the nodes belong. 

\paragraph{Bioinformatic datasets.} D\&D contains graphs of protein structures. A node represents an amino acid and edges are constructed if the distance of two nodes is less than $6\AA$. A label denotes whether a protein is an enzyme or non-enzyme. PROTEINS is a dataset where the nodes are secondary structure elements (SSEs), and there is an edge between two nodes if they are neighbors in the given amino acid sequence or in 3D space. The dataset has 3 discrete labels, representing helixes, sheets or turns. PTC is a dataset containing 344 chemical compounds that reports the carcinogenicity of male and female rats and has 19 discrete labels. NCI1 is a dataset made publicly available by the National Cancer Institute (NCI) and is a subset of balanced datasets containing chemical compounds screened for their ability to suppress or inhibit the growth of a panel of human tumor cell lines; this dataset possesses 37 discrete labels. MUTAG has seven kinds of graphs that are derived from 188 mutagenic aromatic and heteroaromatic nitro compounds. PTC includes 19 discrete labels and reports the carcinogenicity of 344 chemical compounds for male and female rats.

\paragraph{Initial inputs.} The data of the bioinformatic datasets and social network datasets differ in that the nodes in bioinformatics graphs have categorical labels that do not exist in social networks. Thus, the initial node features of the HRN inputs are set to one-hot encodings of the node degrees for social networks and a combination of the one-hot encodings of the degrees and categorical labels for bioinformatic graphs.

\paragraph{Implementation details.} We evaluate the model performance with a 10-fold cross validation setting, where the dataset split is based on the conventionally used training/test splits \cite{zhang2018end,bianchi2020spectral,baek2021accurate}. In addition, we use the 10 percent of the training data as a validation data following the fair comparison setup \cite{errica2020fair}. We use the early stopping criterion, where we stop the training if there is no further improvement on the validation loss during 50 epochs. Furthermore, the maximum number of epochs is set to 500. We then report the average performances on test sets, by performing overall experiments 10 times.
% 小数据集像GIN一样 train 50轮一个epoch
In particular, following the implementation of \cite{xu2019powerful}, we train each epoch with a fixed number of iterations (i.e., 50) for small datasets.
We set the pooling ratio as 25\% in each pooling layer for baselines as previous works \cite{baek2021accurate,bianchi2020spectral}, while our model follows the natural cluster assignments produced by Algorithm 1 with given height 3.
For model configuration, the learning rate is set to $5\times 10^{-4}$, the hidden size is set $\in\{64, 128\}$, the batch size is set $\in\{32, 128\}$, weight decay is set to $1\times 10^{-4}$, and dropout rate is set $\in\{0, 0.5\}$. 
Then we optimize the network with Adam optimizer. For a fair comparison of baselines \cite{lee2019self}, we use the three GCN layers \cite{kipf2017semi} as a message passing function for all models with skip connections, and only change the pooling architecture throughout all models. Because GMT is the most recent work that replicates these popular pooling approaches in previous studies, and we implement SEP-G based on the code of GMT \footnote{\url{https://github.com/JinheonBaek/GMT}}, thus the accuracies of baselines are derived from \cite{baek2021accurate}.

\subsection{Node Classification}

\paragraph{Citation datasets.} We utilize three standard citation network benchmark datasets: Cora, Citeseer and Pubmed \cite{sen2008collective}. In all of these datasets, nodes correspond to documents and edges to (undirected) citations. Node features correspond to elements of a bag-of-words representation of a document. Each node has a class label. In particular, the node features are different among three datasets. Specifically, the input features of Cora and Citeseer are one-hot embedding of words in each document, while the node features of Pubmed are the TF-IDF weighted word vectors. The Cora dataset contains 2708 nodes, 5429 edges, 7 classes and 1433 features per node. The Citeseer dataset contains 3327 nodes, 4732 edges, 6 classes and 3703 features per node. The Pubmed dataset contains 19717 nodes, 44338 edges, 3 classes and 500 features per node.

\paragraph{Implementation details.} We closely follow the transductive experimental setup in \cite{kipf2017semi}. Each class is only allowed 20 nodes for training, which honors the transductive setup, and the training algorithm has access to all of the nodes’ feature vectors. The predictive power of the trained models is evaluated on 1000 test nodes, and we use 500 additional nodes for validation purposes. 
We also use the early stopping criterion, where we stop the training if there is no further improvement on the validation loss during 50 epochs. Furthermore, the maximum number of epochs is set to 1000.
We obtain cluster assignment matrices of each dataset under the guidance of three-dimensional structural entropy, and adopt the first two layers for hierarchical pooling.
For model configuration, the learning rate is set to $0.01$, the hidden size is set $\in\{16, 32, 128, 256\}$, weight decay is set $\in\{0.02, 5\times 10^{-4}\}$, and dropout rate for each layer is set $\in\{0, \dots, 0.9\}$. Finally, we optimize the network with Adam optimizer. 

\paragraph{Fair comparison with SOTA methods.}
To make a fair comparison with S$^2$GC and GCNII, we present the results of the two methods with similar convolutional layers. As shown in Table \ref{tab:depth_sota}, SEP-N obtains competitive results with only 5 GCN layers, which is much less than the requirement of S$^2$GC (16) and GCNII (64 for Cora, 32 for Citeseer, 16 for Pubmed). Furthermore, SEP-N outperforms S$^2$GC and GCNII when putting similar number of convolution layers, which reveals the effectiveness of hierarchical graph pooling in node classification tasks when facing limited computing resources.

\begin{table}[!ht]
\centering
\caption{\small{S$^2$GC and GCNII with similar \#Convs of SEP-N.}}
\label{tab:depth_sota}
\vskip 0.15in
% \resizebox{0.65\textwidth}{!}{%
\begin{tabular}{l|ccccc}
\hline
         & \multicolumn{5}{c}{Model (\#Convs)}                  \\ \hline
         & S$^2$GC(4) & S$^2$GC(8) & GCNII(4) & GCNII(8) & SEP-N(5) \\ \hline
Cora     & 79.8    & 82.2    & 82.6     & 84.2     & \textbf{84.8}     \\
Citeseer & 72.6    & 72.7    & 68.9     & 70.6     & \textbf{72.9}     \\
Pubmed   & 79.2    & 79.7    & 78.8     & 79.3     & \textbf{80.2}     \\ \hline
\end{tabular}%
% }
\vskip -0.1in
\end{table}

\end{document}